\documentclass[12pt,a4paper]{report}
\usepackage{pdfpages}
\usepackage{amsmath}
\usepackage{graphicx}
\usepackage[toc,page]{appendix}
\usepackage[framed,numbered,autolinebreaks,useliterate]{mcode}
\usepackage[lmargin=3.81cm,tmargin=2.54cm,rmargin=2.54cm,bmargin=2.52cm]{geometry}
\pdfoutput=1
\usepackage{mathptmx}

\begin{document}
\includepdf[pages={1}]{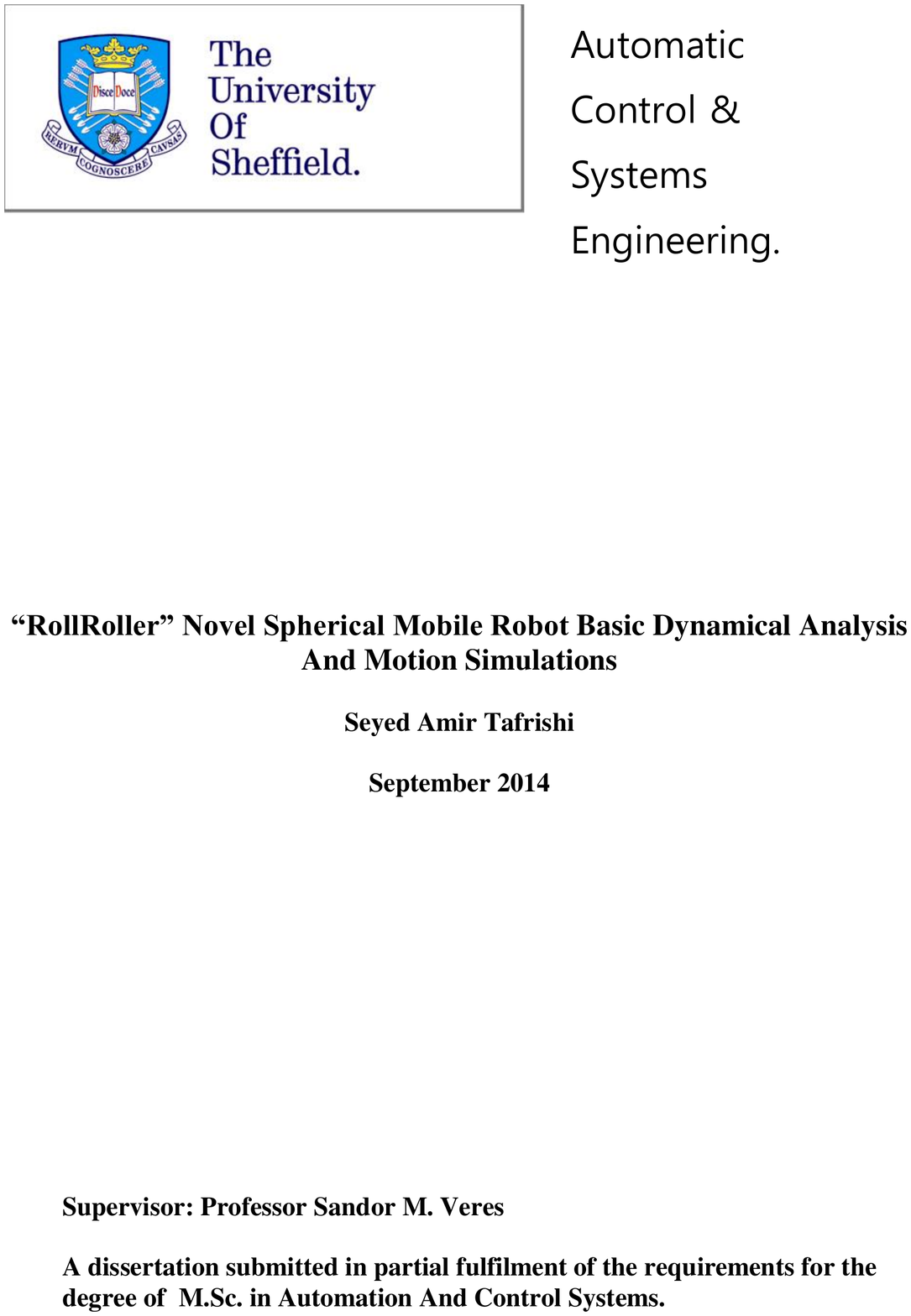}
\pagenumbering{Roman}

\section*{EXECUTIVE SUMMARY}

\subsection*{\emph{INTRODUCTION}}
This project gives novel and different preceptive to unmanned ground vehicles from analyzing unique mechanism under title of spherical mobile robot. In addition, to convince reader about its implementation potential, mechanism of invented omnidirectional air actuated robot ,"RollRoller", checked and then relevant logical algorithms are proposed. Mathematical dynamics were derived to strengthen the model and gain understanding about its morphology to basic motions. Lastly, different type of numerical simulations of robot were created to let investigation about practical design be highly operational for future implementations in exploration and rescue missions.

\subsection*{\emph{AIMS AND OBJECTIVES}}
\textbf{\textit{Aim of Thesis}}

The main goal of this thesis is to understand invented spherical robot. And, investigate theoretical logics such as mathematical derivation of dynamics and simulation to validate "RollRoller" as futuristic alternative operational mobile robot.
\\
\textbf{\textit{Objectives of Thesis}}

\textbf{*}Analyzing "RollRoller" robot's mechanism and required motion algorithms.

\textbf{*} Make comparison of "RollRoller" with other mobile robot types.

\textbf{*} Drawing robot on Solidworks for numerical simulation implementation and possible motion explanations.

\textbf{*} Calculate nonlinear motion (e.g. forward direction) dynamics of robot via Euler–Lagrange method.

\textbf{*} Designing simulation and analyzing the results on Matlab Simulink.

\textbf{*} Importing Solidwork Model and confirming the obtained results on Adams/view program.

\textbf{*} Designing electrical specified "RollRoller" core positioning sensor and control actuators.

\textbf{*} Checking other extra locomotions, particularly circular motion of robot on Simulation.

\subsection*{\emph{ACHIEVEMENTS}}

\textbf{I)} RollRoller's manners and structure were studied and its algorithms were developed.

\textbf{II)} Nonlinear dynamics and state-space model for "RollRoller" basic motion with either friction or without were calculated.

\textbf{III)} Matlab Simulation model designed and results were verified.

\textbf{IV)} "RollRoller" Solidworks model created and imported to Adams/view Numerical Simulation.

\textbf{V)} Adams/view model visualized and results were inspected.

\textbf{VI)} Turning motion were designed on Adams/view simulation and results plotted.

\subsection*{\emph{CONCLUSIONS}}
The proposed novel robot, "RollRoller", performs successfully as expected in all the stages of simulations ( i.e. Matlab Simulink and Adams/view simulation). And obtained results are validating that this spherical mobile robot can be very promising alternative for current unmanned ground vehicles (e.g. Legged, Wheeled and Snake ) in practice. Space investigations, rescue missions and other civilian activities are the possible tasks that RollRoller is able to enroll in near future applications.

\newpage \section*{ABSTRACT}

This paper introduces novel air actuated spherical robot called "RollRoller". The RollRoller robot consists of two essential parts: tubes covered with a shell as a frame and mechanical controlling parts to correspond movements. The RollRoller is proposed to be high potential alternative for exploration and rescue missions robots because robot utlizing its locomotion via all possible deriving methods (gravity, torque and angular momentum forces). In beginning ,  characteristic and role of each of component and features were explained. Next, to determine the uniqueness of this robot the known and other extra possible motions are shown by proposing their own algorithmic movements. To illustrate main motion of this robot was inherent to mathematical models,  the forward direction  dynamical behavior on flat surface was derived. Additionally, Matlab Simulink was used to plot the results to validate the behavior for both fractional and non-fractional terrains. Lastly, after desgining the model of robot in Solidworks Program, Adams/View visualization software  ( the robot simulated form ) was utilized to proof the Matlab Simulink results and to show the more detailed and complete form of locomotion including the forward direction and circular locomotion in proposed robot.

\newpage \section*{ACKNOWLEDGEMENTS}

Initially, i would like to express my sincere gratitude to my project supervisor Prof.Sandor M. Veres about his priceless support in developing my invention as research thesis. His noteworthy advises and given opportunities, helped me to analyze this robot more in advance and also it has given a chance to implement in practice in near future.

In addition, i would like to extend my special appreciation to my B.Sc. supervisor Dr. Larissa Khodadadi, her unforgettable life long guide boosted me in developing my project. And also, her inspiring impact during my B.Sc. studies is main reason in my enhanced creativity and problem solving perspective. Also, Special thanks to all academic members and stuff in University of Sheffield that were helping me during this project. Particularly my M.Sc. Personal Tutor Dr.Jun Liu for his vise advices to let me see the problems more clear.

I dedicate this thesis to my beloved parent and friends specially  Ataback Maleki and Pedram Purali, who have stand beside me during all hardships and always motivated me. Their continues support have helped me mature into person i am today.

\tableofcontents
\listoffigures
\listoftables
\thispagestyle{empty}

\newpage \section*{List of Abbreviation}

\begin{tabular}{ll}
3D & Three Dimensional\tabularnewline
CH$_4$ & Methane \tabularnewline
CM & Center of Mass\tabularnewline
CO$_2$ & Carbon Dioxide\tabularnewline
DOF & Degree Of Freedom\tabularnewline
FC & Fire Circle\tabularnewline
FT & Forwarder Tube\tabularnewline
GB & Gravity Breaker\tabularnewline
GPS & Global Positioning System\tabularnewline
HT & Horizontal Tube\tabularnewline
IC & Integrated Circuit\tabularnewline
IDU & Internal Driving Unit\tabularnewline
IMU & Inertia Measurement Unit\tabularnewline
MD & Main Direction\tabularnewline
MM & Momentum Maker\tabularnewline
NASA & National Aeronautics and Space Administration\tabularnewline
PD & Penetration Depth\tabularnewline
SMR & Spherical Mobile Robot\tabularnewline
TT & Turner Tube\tabularnewline
UGV & Unmanned Ground Vehicle\tabularnewline
VT & Vertical Tube\tabularnewline
ZMP & Zero Moment Point\tabularnewline

\end{tabular}

\chapter{Introduction}

\begin{figure}[ht!]
\centering
\includegraphics[width=100 mm]{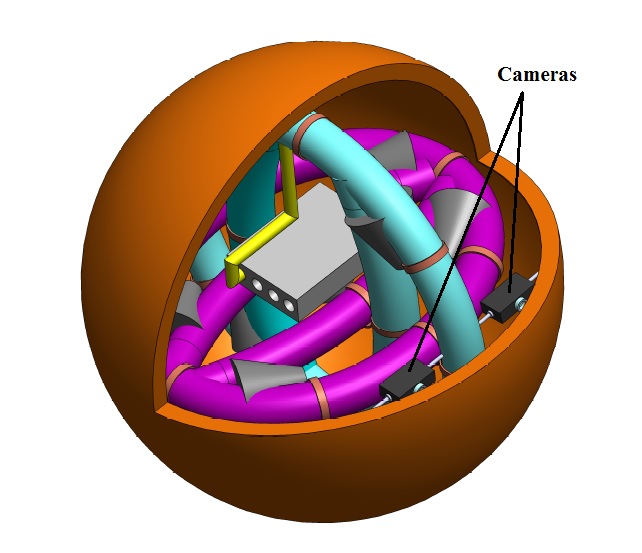}
\caption{RollRoller Robot.}
\end{figure}

Many research areas related to Mobile Robots (e.g. Wheeled, Legged, Bio-inspired and so on) have been created during these centuries  for challenging in certain type of environments with specific pre-defined tasks ,and majority of currently happening researches are focused on proposing techniques for reliable autonomous map building under dynamical environment or designing fast and accurate control methods. However, basically most of these robots are suffering from not being capable to cope with notoriously changeable environment. In this part , firstly the basic three known mobile robots (i.e. Wheeled, Legged, snake robots) will be explained historically and also the pros and cons of each type will be pointed out with noteworthy comparison. Secondly, the main focus will be on spherical robots types and historical background. And in next section, "RollRoller"'s motivation, advantages and disadvantages and also potentiality in near future practical implementation will be explained. Lastly, the achieved aims and objectives with timetable deeds have been written.
\pagenumbering{arabic}
\setcounter{page}{1}
\section{Background and Motivation of Mobile Robots}
\subsection{Wheeled Robots}

The Wheeled Mobile Robot was started to be under the scope seriously by end of 19th \cite{[1]} in which possible wheeled structures proposed and dynamically derived to understand mobility. Next, The Sojourner Rover [2] was landed on Mars by using successful wheeled mechanism, figure 1.2 a shows this robot on mission. Other related studies were taken place during those years [3-5]. However, having fixed mechanism stopped them toward entering to uneven surface and unpredictable environments. The Marsokhod [3] was able only to over come certain roughness proportional to wheel size (figure 1.2 b). Furthermore, likelihood about appearing unknown environment condition was so high such as sand storm, toxic rains and so on so the robots was not able to performance acceptably. Hybtor [6], Spacecat [7] and nanorovers [8-10] were other types with complex control methods that were proposed in which the previous problems were still existed hence they were suitable for places that the specifications and area were deterministic. eventually, latest designed flexible rovers [11-12] still were not suitable to enter places with stochastic atmosphere because the fundamental designs are still untouched and also actuators and motion making wheels were in contact with outside. These were main reasons that prevent this type to enroll in rescue mission and even more the period of involvement of pure wheeled robot in exploration missions in outer surface has got no attraction in these years research ares.
\begin{figure}[ht!]
\centering
\includegraphics[width=50 mm]{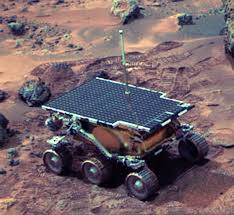} \space a.
\includegraphics[width=40 mm]{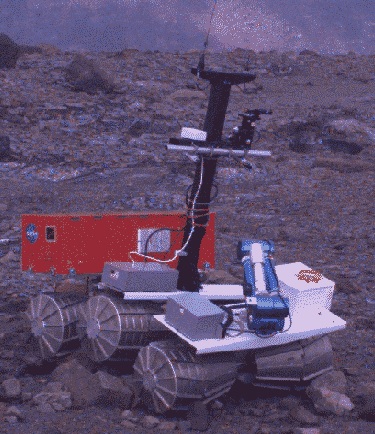} \space b.
\caption{ Wheeled Robots: a.Sojourner b.marsokhod.}
\end{figure}
\subsection{Legged Robots}
Passive Two-Legged robot's dynamics analyzed firstly in 1990 [13] by McGeer. During early 20th human-like legged robots with active joints were designed (figure 1.3 a) but the problems about controlling appeared because of singularities and complex dynamics [14-15].Authors of [16-18] proposed using advanced control methods for having more flexibility in humanoid robots , although there were still challenges that made these robots unsuitable for any exploration and rescue mission activities. In other words, legged robots are not able to travel through traverse terrain, even by using Zero Moment Point (ZMP) [19]. The speed of these kind is considerably low and also having joints and motion making mechanism in outside the robot prevent them for entering hazardous areas. Most of these kinds are suitable for deterministic and ordinary activities such as society guide and health services. 
\begin{figure}[ht!]
\centering
\includegraphics[width=27 mm]{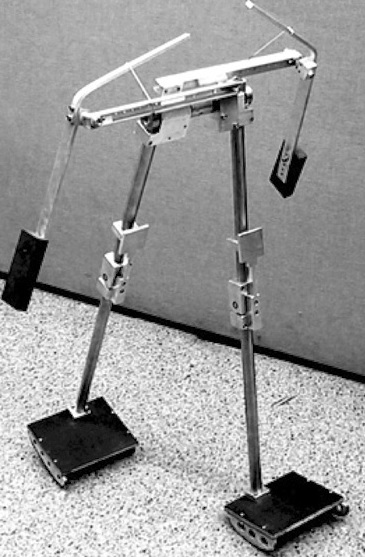} \space a
\includegraphics[width=42 mm]{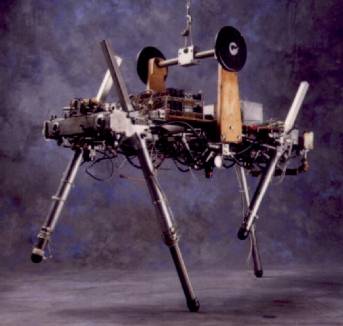} \space b
\includegraphics[width=56 mm]{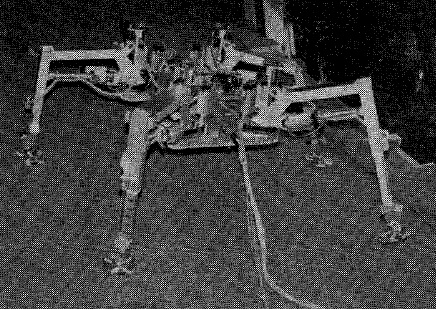} \space c
\\
\includegraphics[width=47 mm]{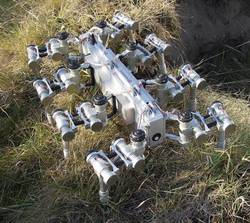} \space d
\includegraphics[width=60 mm]{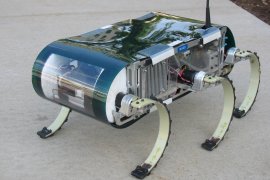} \space e

\caption{ Legged robots: (a).Two-legged (b).MIT's work (c).TITAN VII (d).Rhex (e).SCORPION}
\end{figure}

On the other hand, MacGee as pioneer at multi-legged robots, formulated them mathematically [20].Simultaneously the scholars at MIT started working these type because of their extra Degree of freedom (DOF)[21] which results a practical developments such as figure 1.3 b. Multi-legged robots were more stable and tolerant toward hilly surfaces by comparison with two-legged robots but logically more joints led more complex dynamics and sophisticated manners of robot. As consequence , they were requiring advance control methods with high speed calculation time which still is under development within these years. TITAN VII [22] and Rhex [23] figure(1.3 c-d) also were other developed multi-legged robots with their own benefits as alternative actuation mechanisms for this kind. There were stability test on SCORPION ( 8 legged mobile robot) when it loses its two legs  [24] but still it was lack of practice for real world. As a result, Legged Robots are commonly known for their extra DOF but by getting involve in generic places, their legs may stock or controllers may face ambiguity or serious errors toward unpredicted changes. However, combination of this kind with wheeled structure is newly used as convenient practical deeds, as an example NASA's "Curiosity" robot sent to mars for doing geological sampling.

\subsection{Snake Robots}
Biologically inspired snake robots studied and designed in 1993 by Hirose (figure 1.4 a) [25]. Although, he formulated the ground friction, temperature and snake shape in his dynamics, the robot was able to do only lateral undulation. Furthermore, having complex dynamics and environmentally dependent characteristics made these type less attractive, as a result it found more practical in activities like pipeline inspections [26]. Till now different snake robots such as ACM-R3 [27], Perambulator-II [28] and other models [29-30] have been theoretically evaluated and developed. Nevertheless , they were not operational for any serious activities since they were responding slow  and sensitive to outside world. Scholars in [31] designed snake robot for rescue mission but research was on theoretical base. Overall, although snake robots have joints outside, there is way to cover it up. But again not having enough space for extra instruments, not having jumping capability and high likelihood of stocking/locking under crashed buildings stops them from being superior choice. Because of that this type is still under development from mechanical and controller design aspects.
\begin{figure}[ht!]
\centering
\includegraphics[width=45 mm]{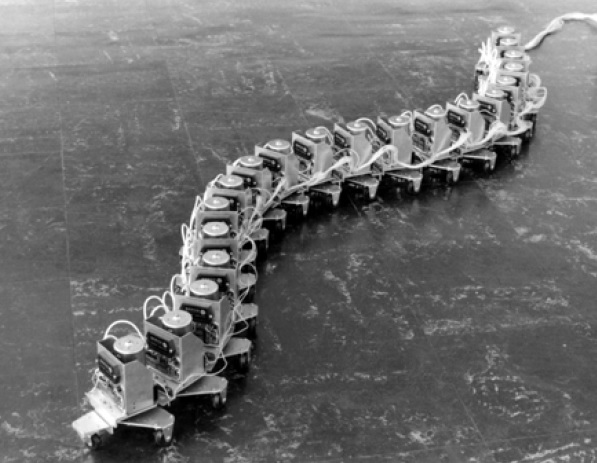} \space a
\includegraphics[width=47 mm]{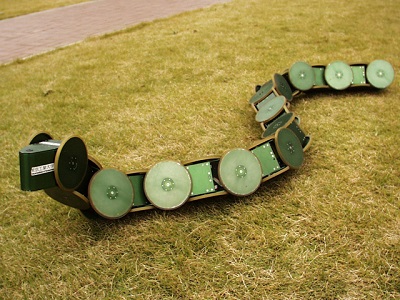} \space b
\includegraphics[width=40 mm]{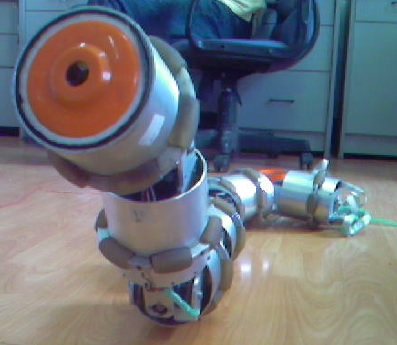} c
\caption{ Snake robots: (a).Hirose's robot,(b).ACM-R3 ,(c).Perambulator-II}
\end{figure}
\section{Spherical Robots History And Characteristic}
Spherical Robots are revolutionary mobile robots because of their symmetrical surface by comparison with other types. In this type all required locomotion motions and other instruments are just wrapped inside spherical shell. Spherical Mobile Robots (SMRs) can move in all directions, swing , spin proportionally or individually and even can jump in all kind of unknown terrains ( all proposed mechanism types are shown in figure 1.5). As result, SMRs are being considered to enroll in serious and hazardous operations such as rescue or exploration missions . Not only this kind is able to show phenomenal potential in serious tasks but also can have active role in industrial and social services. On the other hand, sophistication of actuation mechanism in SMR's  and limitation of activities that are simple in rest of mobile robots particularly in practical areas led this model to not to be as attractive as them. About driving mechanism spherical mobile robots can be divided into three categories as general: Torque , Gravity and Angular Momentum Driven.
\subsection{Torque Driven}
In Torque Driven method , sphere gets its motivation force by using motors reaction force directly connected to spherical shell inner surface. This concept firstly was implemented and analyzed about mono-directional dynamics  by Halme et al in 1996 as figure 1.5 [32]. 

\begin{figure}[ht!]
\centering
\includegraphics[width=140mm]{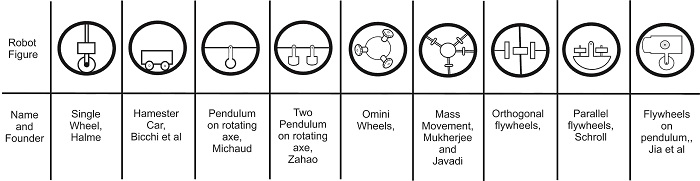}
\caption{Past proposed spherical robots.}
\end{figure}

The figure 1.6.a shows the sphere with centralized mass which can be considered as general definition, and also the spherical shell mass by comparison of total robot mass is neglected. Figure 1.6.b can be showed as unique form of this kind. The center of mass in this method shifted with $d$ distance from sphere center and outer shell is connected to this CM.In other words, it can be imagined as pendulum driven mechanism. By giving angular change via "u" torque in counter-clockwise direction, connected centralized mass object moves. Then, the robot gains opposed torque from it so makes total sphere to displace along terrain. This motion maximum speed can be estimated when the pendulum moves the CM to around the horizontal line region. Figure 1.7 illustrates the power of sphere motion respect to CM. And, Distance from ground is defining the smallest angular change with relativity to speed hence as much as the center of mass moves beneath the center and reaches to ground, although utilizing the accurate motions become possible, sphere is not capable of having high torque reaction during locomotion. The same manner happens when the center of mass pass and goes upper than sphere's center, this may achieve the highest possible torque but if it is using pendulum driving mechanism for instance, system will be unstable. As result, maximum safe torque is achievable with only locating at center as figure 1.7 c, however due to relative connection between center of sphere, beam and mass it is not practically achievable. 

The approximated travel distance with its relation to sphere radiance and angle change ($\theta$) derived from geometric characteristic (figure 1.7 c).
\begin{equation}
d(rad)=\frac{2\pi R}{360} \theta
\end{equation}
Kinetic relativity due to angular change in pendulum shows that as distance from vertical line's CM increases , the travel raises proportional to velocity.
\begin{figure}[ht!]
\centering
\includegraphics[width=68mm]{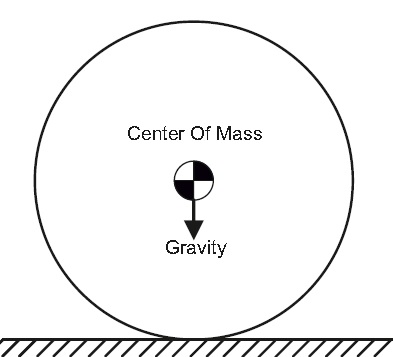} \space a
\includegraphics[width=68mm]{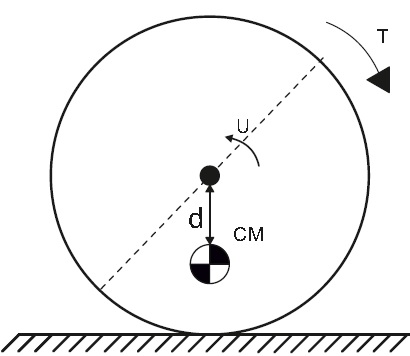} \space b
\caption{Unique form of torque driven method.} 
\end{figure}

Single wheel structure was other used mechanism in which had only one degree of freedom (DOF) and internal driving unit was unstable toward overcoming serious uphill or ramp.Bacchi et al introduced wheeled car placed in sphere changing the center of mass by displacing of the car [33]. Although "SPHERICLE" had simple IDU structure, it was occupying the whole sphere and also were unstable for sudden uneven terrains. Then, a pendulum driven toy ball proposed and analyzed from physiological and safety aspects [34] (no physical stability in motor high speed of motors). RoBall [35] and Volvolt [36] as latest models of torque driven IDU were still not able to have perfect omni-directional movement (i.e. errors in circular and slide motions).\\

\begin{figure}[ht!]
\centering
\includegraphics[width=65mm]{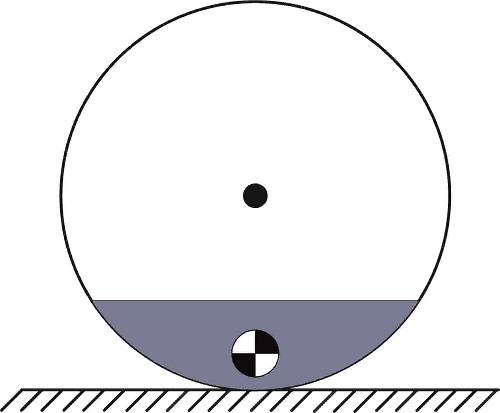} \space a
\includegraphics[width=65mm]{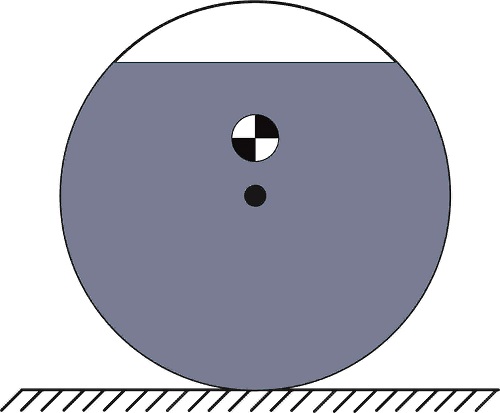} \space b
\includegraphics[width=65mm]{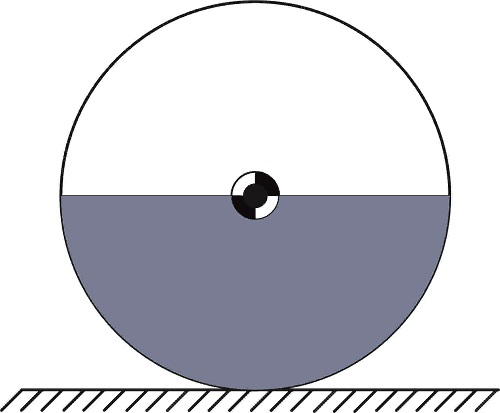} \space c
\includegraphics[width=65mm]{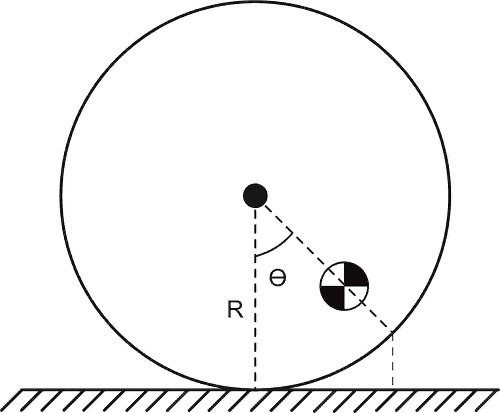} \space d
\caption{ The torque limitations [42]: (a). Minimum torque (b). Maximum unstable toruqe (c).Maximum stable torque (d). The travel distance relation.}
\end{figure}

\subsection{Gravity Driven}
In the Gravity Driven IDU, position of robot changes with respect to center of mass displacement(figure 1.8). However, there is no hard relation with outer shell in this motion typology. The gravitational acceleration takes place by moving center of mass with x distance inside sphere to equalize the center of mass.  Mukherejee et al developed  "Spherobot" as pioneer in this field in 1999, moving masses on four spokes connected from center of sphere to surface of shell as mechanism  [37], to generate required motion. Javadi did further analysis on this concept about dynamic and trajectory planning [38-39]. Nevertheless, Gravity driven IDU alone were suffering from lack of continues acceleration and high velocities and they weren't able to climb ramps or rough terrains as well. In contrast with torque driven this locomotion can have more accurate and slow motions along the path. 
\begin{figure}[ht!]
\centering
\includegraphics[width=70mm]{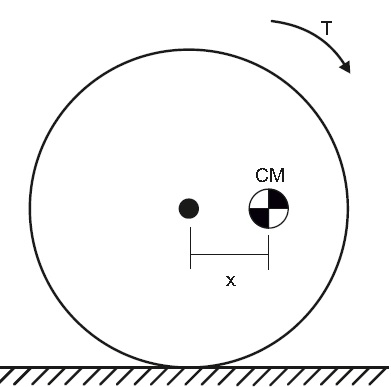}
\caption{Unique form of gravity driven method.}
\end{figure}
\subsection{Angular Momentum Driven}
These SMR's motion torque produced by conserving the angular momentum created in IDU. This locomotion can be produced via either of driving methods. Bhattacharya et al proposed two perpendicular motors inside sphere in 2000 [40]. Motors with creating individual or co-operative momentum generate total movement for spherical robot but having complex motion mechanics and dependency on rotor's speed decrease robot's efficiency. In 2009, scholars developed control method for angular momentum force by using the pendulum structure with detached relation of spherical shell [41]. Also, The result was controlled rotation with angular momentum of IDU. However, generally this type of driving mechanism is not suitable to be main stream for robot's motion and it may find more practical usage in outer space or places that gravity and frictional reaction-torque are not efficient anymore. Having considerably small magnitude of displacement force , high sensitivity of disturbance and noise makes them weak alternative from other driving IDUs. From then on, many researches have been taken place to combine angular momentum with other methods. For example, Schroll designed and built "Gyrosphere robot" that was able to use  gravitational and angular momentum forces to have different motions in 2008 [42]. Another combination of direct driving ( Torque Reaction) and angular momentum force was developed by Jia et al in 2009 [43]. These latest SMRs' control methods and ability to behave perfectly in interchangeable environment were considerably low yet again. Even more, they mostly were lack of jumping capability and enough space to other required instruments.
\subsection{Other Types of Spherical Robots}
In 2005 a deformable SMR was studied by Sugiyama to pass crawl rough terrains and jump as figure 1.9 [44]. The deformable SMR consisted of eight wire connected to elastic surface in which changing length of each of these wires reform the SMR. Despite these capabilities, the robot was extremely sensitive to be unstable due to complex mechanism. Consequently, robot still is under development for practical utilizations.
\begin{figure}[ht!]
\centering
\includegraphics[width=140mm]{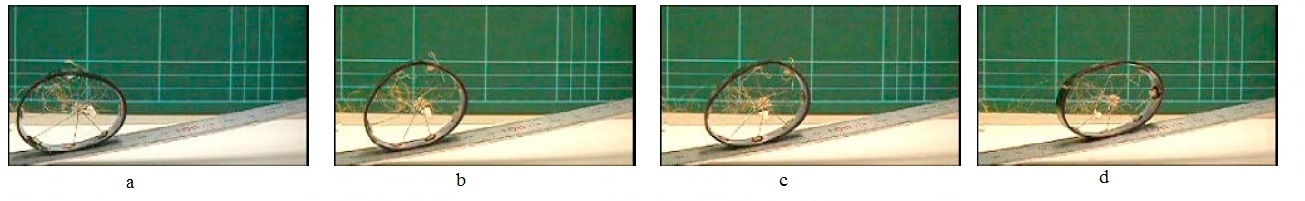}
\includegraphics[width=140mm]{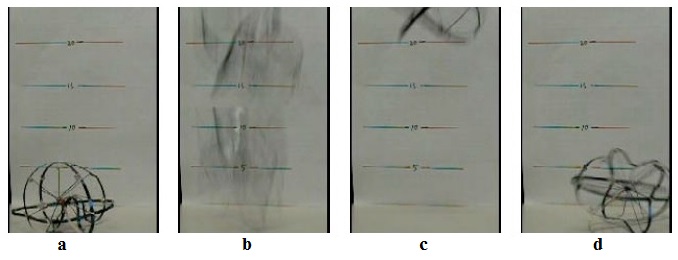}
\caption{The deformobale robot's ramp climb and jump [44].}
\end{figure}

\begin{figure}[ht!]
\centering
\includegraphics[width=140mm]{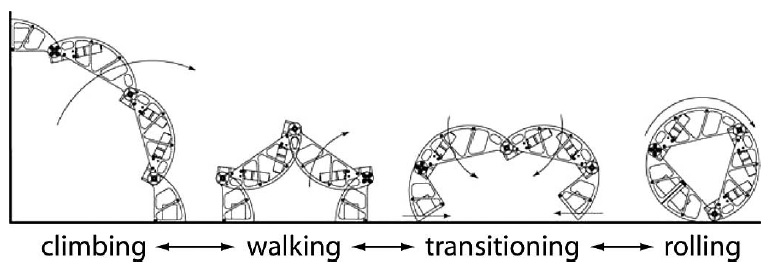}
\caption{The Hybrid SMR [45].}
\end{figure}
Phipps et al. [45] designed rolling disk-like biped hybrid robot in 2008 which was able to walk, climb and roll as figure 1.10. Efficiency of rolling in robot was validated as highest one from walking but the dynamical motion and control methods didn't designed yet in which it may contain more complexity by comparison with fixed SMRs.
scientists have been working on a new deformable spherical rover actuating with heat input to sphere surface(figure 1.11) [46]. However; Electroactive Elastomeric Actuators make system unsuitable in today tasks because this phenomenon requiring extra heating mechanism with huge amount of energy requirements to deform the sphere surface. Moreover, small error in system controlling method may cause inevitable damages to sphere surface hence the robots dynamics changes relative to heat, and that may result crash down in operating system.

\begin{figure}[ht!]
\centering
\includegraphics[width=140mm]{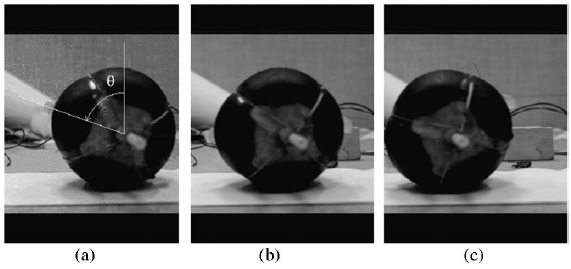}
\caption{Electroactive Elastomeric Deformable Rover [46].}
\end{figure}

\section{Motivation of RollRoller Invention}
In this study, a novel spherical mobile robot “RollRoller” with practical capabilities with comparing with other spherical robots is presented. The proposed robot, shown in Fig.1.1, can make its momentum force by all three methods (i.e. pendulum driven, angular and torque movements) functionally. In my invention number of DOF increases to four to advance the mobility (i.e.  Forward/Backward, lateral  and orientation motions) and adjusting significant feature as jump activity. RollRoller take its forward momentum force from center of mass change mainly via displacement of copper orbs inside robot plus the torque reaction force so it can approximately considered as pendulum-driven spherical robot despite the more freedom for complex movements (mixture of diverse motivation forces) which for most of previous spherical robots only two driving methods ( either angular-gravity or angular-torque) were achievable. Also, it can have extra equipments like manipulator arm, telescope, digger, or other research instruments because of isolated mechanism inside the unshared pipes.  This designed robot could show incredible function in exploration projects or as rescue robot in natural disastrous events.

Although, the novel proposed SMR (RollRoller) overcomes many disadvantages of other mobile robots that explained, it has many practical and simulation designing hardship to overcome. For instance bellow events were some of hardships that can be notified on this dissertation:

\textbf{I)} Proposing the movements with their own algorithms particular for RollRoller, were challenging procedure since it has to be imagined with including all environment effects robot itself together (done before the course). however, eventually the required logical and controllable algorithm that will be unique and simple has been found and explained.

\textbf{II)} Another hardship was the calculation of nonholonomic nonlinear dynamics in spherical robot in which after studying many papers and doing analysis,  most convenient and useful novel dynamics derived for RollRoller IDU mechanism that let us estimate the linearized form for designing required controller in future researches. Additionally, more general form generated with adding frictional effects to robot.

\textbf{III)} Designing tubes' and cores' size, mass and materials type were another serious challenge since one of most essential decisive point was its  characteristics making robot logical and possible to be implemented. Hence, first the importance of each parameters defined and then designed relative to priority. For instance, changing the size of robot doesn't have direct relation with IDU mechanis. Nevertheless, at end all the input data's and physical definitions made to be practically applicable in this dissertation.

\textbf{IV})Lastly, air input force that makes the corresponded core to move, was problematic to design. Particularly, during the designing the numerical simulation Adams/view due to complexity of nonlinear air dynamics it was estimated and designed with solid object (pendulum pusher) instead. However, this phenomenon didn't have any negative contribution on changing the expected results.

This paper is represented in three main sections. Section 2 involves physical feature of RollRoller. The following subsections are frame of robot, explanation of mechanical instruments, Movement algorithms and explaining electrical sensors. Derived Forward direction dynamics with and without friction are demonstrated in Section 3. Section 4 is written about different simulation methods(Matlab Simulink and Adams/view numerical simulation). 

\section{Application Opportunities and Potentials}
\subsection{Rescue} 

\textbf{1.Fire Rescue Robot:}  "RollRoller" robot with a covered fire redundant alloy around the spherical shell can involve in rescue missions as fire extinguisher since it has got considerable space inside itself that let robot to have other instruments without interaction with SMR's main mechanism. Furthermore , this type is more flexible, accurate and faster than human being fire fighter. For example, it can enter to burning crashed  building , analyze surrounding by sensors , find the victims , check the area temperature , find all the possible safe paths with low $CO_2$ and $CH_4$ level , reach place without affection from fire temperature , use fire extinguisher nozzle to clear path and let victims leave building successfully or send signal to human fire fighters to take the situation under control. Additional way is passive enrollment as assistant rescue robot to help fire fighters in caring heavy and sensitive materials in their coverage. Having fast response to prevent any damage to fire fighter or victim is another way to be utilized it as active assistant rescue robot.

\textbf{2.Nuclear or Hazardous Area Inspector:} First and foremost, safety of pipelines and tanks in sensitive plants such as nuclear plant is crucial. The Fukushima nuclear accident in Japan showed the importance of this phenomenon in which wrong display on air articulator pipes made the Fukushima district inhabitable. As a solution, RollRoller can inspect leakage or damage of pipelines and tanks. In next stage, By advancing the robot to reparteeing items , simultaneous sense and act may be implemented to repair. Also, the same as Fire rescue activity this robot can work actively as rescue mobile robot. Proposed SMR is stronger  than other type of mobile robots (e.g. Legged, Wheeled and snake) since initially all the actuation and mechanical features are warped in sphere. Secondly, SMRs are more environment friendly because of their symmetric shape.\\

\subsection{Exploration}
\textbf{1.Autonomous Ground Space Exploration Vehicle:} Space mobile robots are designed for planetary exploration and collecting sample during space mission. "RollRoller" is revolutionary type for this task since past robots (e.g. Mars Rover, Curiosity) were only used in known and deterministic environment but "RollRoller" doesn't require satellites observation about weather, doesn't need 24 hours live observation to respond any unknown event by operators, doesn't required to have very complex control agents or huge structure to prevent it from collapsing.As a result , with medium scale and low budget sending an mobile robot explorer to any planet wont be impossible anymore.Lastly, first polar explorer spherical robot "tumbleweed rover"  was proposed by NASA [47] which shows it's potential for future investigations.

\textbf{2. Space Repairer:} Because "RollRoller" can make Angular momentum force, Torque reaction force and combination of those, it will be practical for outer space activities without gravity. Consequently, non-gravitational places like space stations or satellites will be able to use this type to repair their facilities without sending an astronaut directly.

\textbf{3. Mobile Sailing Robot:} Due to spherical shell formation, vehicle is able to move in both ground and water. However, robot has to have extra equipments to have applicable motion in water.

\subsection{Industrial}

\textbf{1.Pipelines Inspector Robot:} Gas and water pipelines are spread around cities. Detecting the problematic pipes is important challenge to cut extra costs from changing wrong pipe. By including the "RollRoller" in this inspection activities through pipes, robot can detect problematic part easily. 

\textbf{2.Dam Inspector:} Other similar usage as pipelines, is dam observation and maintenance. Dam's control gates and power generators are those parts that not only is not reachable easily by human being but also requires serious inspections about preventing possible crash down in electrical generation process.Generally this robot can enter not human friendly places then inspect , analyze , understand problem, send requiring information to operators and lastly do maintenance on broken or problematic part.

\textbf{3.Agriculture:} In agriculture sector identifying bugs, spraying, removing waste plants are time consuming jobs. Special designed "RollRoller" for farming industry moving through farms , spraying required places to prevent bugs or using specific instrument to take out the waste plants. As consequence, this phenomenon will increase quality of farming products and at same time will raise the income for farmers.

\textbf{4.Building's Painter Robot:} if a paint caring spherical layer placed on robot because of jumping and other coordinative movements on robot, it can perfectly jump and use wall reflected action. This process can be continued and it will paint whole room.

\subsection{Social Involvement}
\textbf{1. Social Guide:} Assisting people in social places is another noteworthy task. For instance , in restaurant assisting costumers to their places , taking orders and giving payment bills, will be suitable for deeds with less time-consumption and better service quality in tasks instead of human waiters hence considerably it will increase the income of business owners and ease the employees' and costumers' job. Furthermore,in contrast with other robots if any unknown object suddenly take the robot out of its task trajectory , without taking any harm to its IDU ,"RollRoller" can return without any sophisticated feedback control algorithms in which it required mostly for joined robots.

\textbf{2. Children Toy:} "RollRoller" doesn't contain any harmful instruments outside the shell. And also this robot via camera,  Inertia Measurement Unit (IMU) and Global Positioning System(GPS) , parents easily will be able to track their children and check them in anytime. Moreover, with using defensive algorithms it will able to protect children from strangers.\\

\section{Aims and Objectives}
\subsection{Aims}
The project aims to analyze novel spherical mobile robot (RollRoller) in which will have futuristic usage in both academic and industrial areas. The proposed robot will be able to overcome many problems by comparison with its own kind and also is a superior alternative for other Unmanned ground Vehicle types (e.g. Humanoid, wheeled and bio-inspired robot). In this project the robot's derived dynamics are going to be analyzed to validate motion mechanisms and gain understanding of complex behaviors. In the last phase of project suitable actuators and sensors will be designed for RollRoller.
\subsection{Objectives}

\textit{1.}	Developing novel RollRoller robot’s Mechanism.\\
\textit{2.}	Compare previous proposed robots. \\
\textit{3.}	Drawing robot on Solidworks for numerical simulation implementation and possible motion explanations.\\
\textit{4.}	Calculate motion (e.g. forward direction) dynamics of robot.\\
\textit{5.}	Simulate and analyze the results on Matlab Simulink.\\
\textit{6.}	Validate the results by designing the robot on Adams/view.\\
\textit{7.} Design actuator and sensors if the logical and acceptable results are obtained.\\
\textit{8.}	Analyze other extra locomotions particularly circular motion of robot.\\

\subsection{Project Management}

\begin{figure}[ht!]
\centering
\includegraphics[width=150mm]{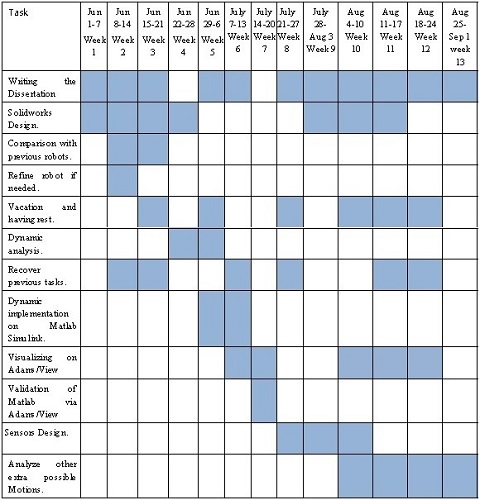}
\centering 
\caption{The Gantt Chart}
\end{figure}
\pagebreak

\chapter{RollRoller Features and Motion Analysis}
\begin{figure}[ht!]
\centering
\includegraphics[width=120mm]{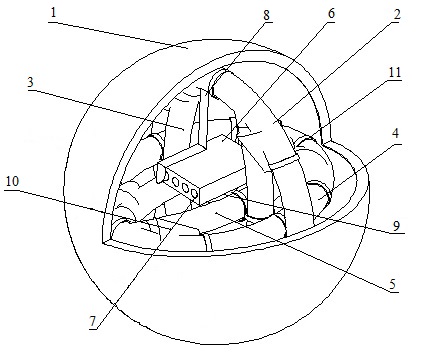}
\caption{RollRoller Structure: 1.Shell, 2.Forward Tube (MM),3.Forward Tube (GB), 4.Turner Tube (MM), 5.Turner Tube(GB), 6.Control Box, 7.Pneumatic motors output, 8.Predefined air injection points, 9.Pressurized air Tank, 10.Control gate, 11.Position sensors.
}
\end{figure}
\section{Physical Features}	

\subsection{Frame of Robot}

RollRoller is covered by plastic transparent shell like Figure 2.2 (i.e., 'a').Then, it is used pipe-like objects for producing movement force in isolation with other equipments as frame and this specification let us to reach flexible motion mechanism. These tubes are made of nylon 12 with high density. Therefore, stamina opposed to unpredicted vibration is vanished and electrical or electromagnetic distortion is impenetrable, also the used nylon 12 characteristic is demonstrated on Table 2.1. The material type of tubes playing crucial role on manner of robot's displacement. For example, if the friction between the core and tube be extremely high , the robot wont be able to reflex to upcoming control instructions.
\begin{table}[ht!]
\begin{center}
\begin{tabular}{|c|c|c|c|c|c|c|c|c|c|}
\hline 
Property & Value\tabularnewline
\hline 
Density  & 1.01 $g/cm^3$\tabularnewline
\hline 
Surface Hardness & RR105 \tabularnewline
\hline 
Tensile Strength & 7251.89 psi (50 MPA) \tabularnewline
\hline 
Flexural Modulus & 1.4 GPa\tabularnewline
\hline 
Notched Izod & 0.06 KJ/m\tabularnewline
\hline 
Elongation at Break  &  200 \% \tabularnewline
\hline 
Strain at Yield & 6 \% \tabularnewline
\hline 
Maximum Operating Temp. &  158$^o$F\tabularnewline
\hline 
Dielectric Strength & 60 MV/m\tabularnewline
\hline 
Heat Distortion Temperature @ 0.45 MPA & 302$^o$F\tabularnewline
\hline 
Material Drying & 2 hours @ 194$^o$F\tabularnewline
\hline 
\end{tabular}
\end{center}
\centering
\caption{CHARACTERISTIC OF CONSIDERED NYLON 12 MATERIAL.}
\label{tab:1}
\end{table}

In proposed robot, two groups of these objects are placed. Main group is called Forwarder Tube (FT) or Vertical Tube (VT) in static position, placed on perpendicular to ground. Next Turner Tube (TT) or Horizontal Tube (HT) in static position, is located perpendicular to the FT. For preventing interaction of two objects with each other, the structure of TT is made with specific functional change. Accordingly, FT frame is circular. And, Turner Tube is in elliptic model, figure 2.2 (a') shows this difference. Furthermore, containing pipes in each tube called as their tasks in machine. For example, the pipes in shape of half-circle called 'Momentum Maker' (MM) line, the majority of motivation force for RollRoller is produced by this type due to alternation of mass and angular momentum force. 'Gravity Breaker' (GB) also is straight path with curve deformation. GB are for making motion more smooth and accurate, and also they are able to let robot have more active motions via including control gates. In addition, The jumping action happens by using GB pipes after getting required speed since inside the GB pipes the loss of energy is becoming minimum.

For each tube, ball-shaped core with copper alloy on surface with 1/3 of complete robot mass is placed. By displacement of these cores inside tubes, the required motion is reachable. However, because there is structural difference inside the HT tube , to make both Tubes' affection relatively same about the gravitational issues, HT's core should be heavier than the VT's core ($m_{HT c} = +2.5$-$10 \% m_{VT c}$ ).

\begin{figure}[ht!]
\centering
\includegraphics[width=70mm]{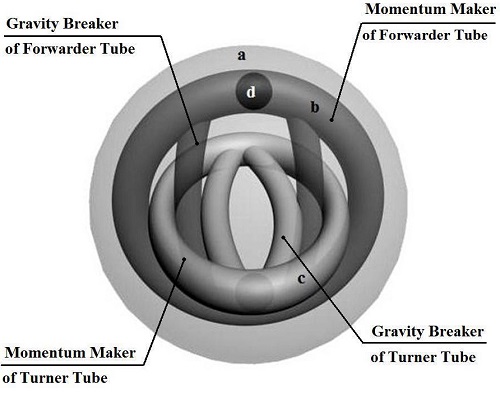}
(a')
\includegraphics[width=60mm]{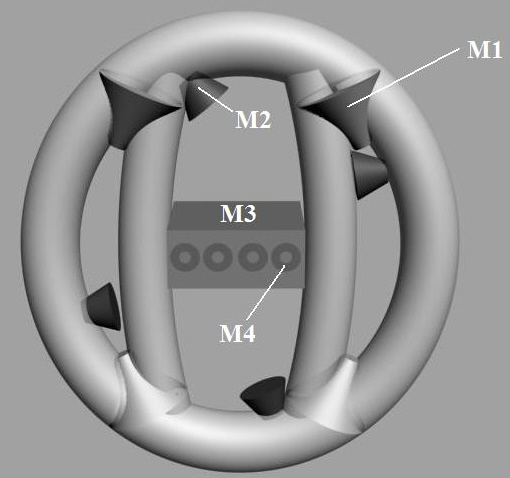}
(b')
\includegraphics[width=60mm]{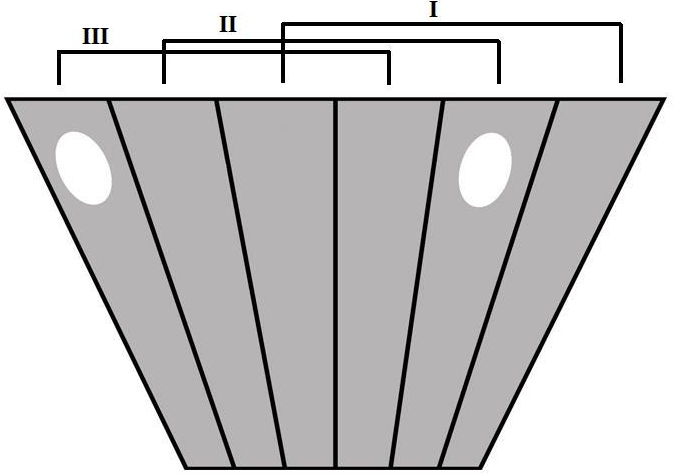}
(c')
\caption{ (a'). Specified parts: a. Spherical Shell, b. Forwarder tube, c. Turner tube. d. Copper core. (b').Mechanical equipments: M1. Control Gate, M2. Input hole for pressurized air, M3. Motor Box, M4.Output of pressurized air. (c'). Flattened control gate.
}
\end{figure}	
\subsection{Mechanical Instruments}	
For giving corresponded force to copper core, pressurized air is produced by pneumatic cylinder joined to stepper motor which total part defines the inside driving unit (IDU). The produced air force enters the tubes with steady magnitude under ideal condition since there is no leakage and always pressure and air quality inside the tube stay constant. To keep the level of air force constant, the air tank is used to increase/decrease pressure inside tube when any leakage sensed. However, another main goal is keeping cores in certain speed hence air actuated force can be vary proportionally. In Figure 2.2 (b'.M3), related actuators are placed inside box and outputs located on box in SMR. These systems can perform in either suction or injection mode. For each FT and TT two motors and two cylindrical pneumatic actuators separately are placed in motor boxes. Each of boxes contain four manual input holes which connected to specific places on tube as seen in picture. This overall system has got capability of controlling the copper bar in majority of controlled algorithms . Control gates are located in predefined accurate places around the tube for manipulating the path of copper core via opening and closing, to act in one of the either functions.

Figure 2.2 (c') shows the flatten form of pyramid control gates locating in interface of MM with GB pipes. During function I activation both pipes are closed so any air movement and possibility of entering the core to other pipes' area is avoided. Function II and III keep an either paths (e.g., Momentum maker) open or closed. Not only control gate behave proportional to place of core and used for changing circulation of air inputs to tubes but also is playing main role in creating the reaction torque in RollRoller, this specification will be explained in jumping motion later on. Particularly, when the core reaches to nearby of gate with constant speed, the gate senses the place of it via position sensors and then form itself relative to simulation.
\section{Movements Algorithm}
\subsection{Forward Locomotion}
\begin{figure}[ht!]
\centering
\includegraphics[width=140 mm]{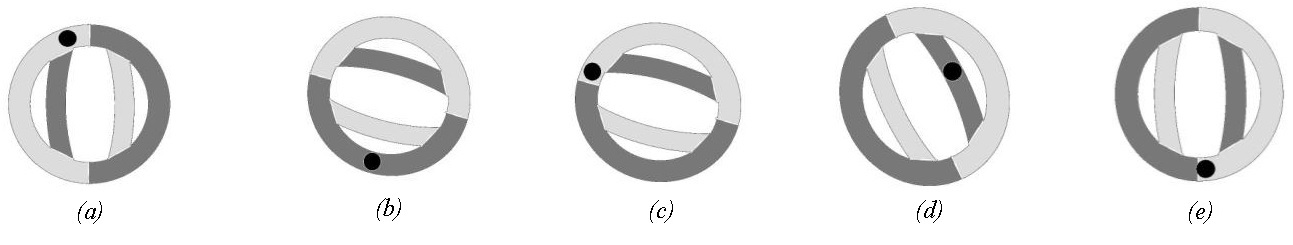}
\caption{Rolling motion model along O-y.}
\end{figure}

This locomotion section is just providing movement in forward direction (figure 2.3) in flat terrain without any damp or obstacle. As shown in Figure 2.4, overall tube is divided to subparts. $\alpha$ part is the beginner of motion and next process is belonging to other part ($\beta$). This reputation continues to keep robot direct line path. Pressurized air input and gate controls are named, Pn and Gn respectively. RollRoller's motion speed is relative to gravity, input air power and pressure, core speed and train type. This motion method is mostly perfect for having steady movements since the speed and displacement of sphere is relatively can be created smooth and stable way due to involvement of gravity  and angular momentum driven method (more advance motions is achievable by involving torque reaction).
\begin{figure}[ht!]
\centering
\includegraphics[width=70 mm]{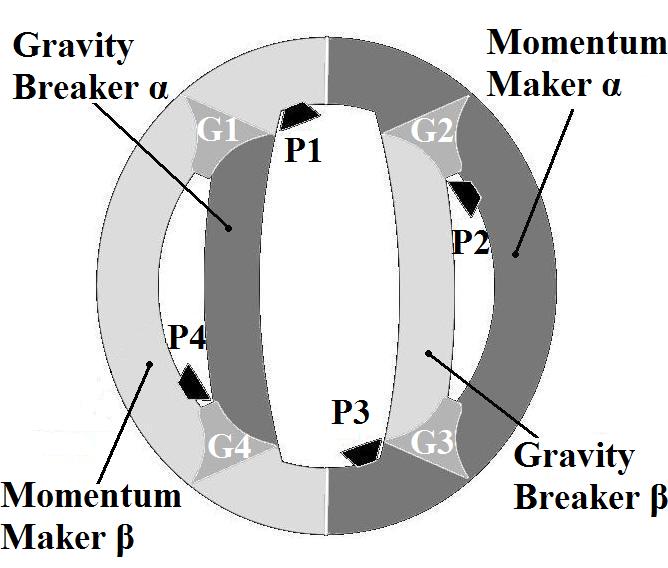}
\caption{The division of tube.}
\end{figure}
\\
The Algorithm of locomotion is as follows:

\textbf{Step 1}: Initialize the location of core inside Forwarder Tube ($X= \alpha$).

\textbf{Step 2}:	Move the core to corresponded place.

\textbf{Step 3}:	Pass through the core to Momentum Maker (MM) X.

\textbf{Step 4}:	Lead the core to the Gravity Breaker (GB) X.

\textbf{Step 5}: if $X= \alpha$ change it to $\beta$. Otherwise, change it to $\alpha$.

\textbf{Step 6}: Go back to Step 2.

The control gates and input pressurized repulsive air from ports are working in cooperative way with each other in this process of motion in which for each cycle of movement, the role of each component is demonstrated on Table 2.1.
\begin{table}[htb]
\begin{center}
\begin{tabular}{|c|c|c|c|c|c|c|c|c|c|}
\hline 
Cycle  & \multicolumn{1}{c}{} & \multicolumn{1}{c}{Control } & \multicolumn{1}{c}{Gates} &  & \multicolumn{1}{c}{} & \multicolumn{1}{c}{Output} & \multicolumn{1}{c}{Ports} &  & Core \tabularnewline
\cline{2-9} 
Mode & G1 & G2 & G3 & G4 & P1 & P2 & P3 & P4 & Location\tabularnewline
\hline 
a & I & II & II & I & I & S & O & O & MM\tabularnewline
\hline 
b & I & I & II & II & O & I & S & O & MM\tabularnewline
\hline 
c & II & I & I & II & S & O & I & O & MM\tabularnewline
\hline 
d & II & II & I & I & S & O & I & O & GB\tabularnewline
\hline 
e & II & I & I & II & I & T & S & T & EP\tabularnewline
\hline 
\end{tabular}
\end{center}
\centering O= Off, S=Suction in tube, I=Injection in tube, T= Injection/Suction in tube via tank , Equilibrium point=EP, GB β= Gravity Breaker β, MM β=Momentum Maker β, GB α = Gravity Breaker α,  MM α = Momentum maker α.
\caption{FUNCTIONS OF FORWARD MOVEMENT}
\label{tab:1}
\end{table}

Avoiding any instability while core is climbing up the sphere to change center of mass is strongest point of this algorithm and robot, in which almost was such a impossible for most of previous robots. Even more algorithmic steps is in a way that failure in control method while there is an air inputs change requirement, doesn't reflects sudden fluctuations in total robot's displacement. Lastly, one cycle delay due to friction factors or imperfect input may just prevent the robot to enter next cycle mode. However, in this case IMU and gravitational sensors would be involved and let robot to repeat cycle until the required position is achieved by SMR.
\subsection{Circular Turning}
\begin{figure}[ht!]
\centering
\includegraphics[width=140 mm]{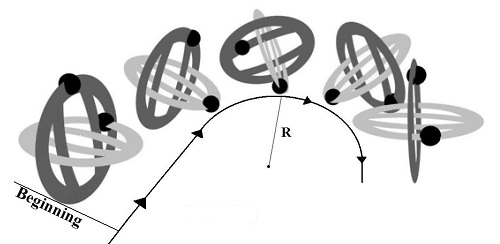}
\caption{Cycle of circular locomotion.}
\end{figure}
In this particular movement, the proposed robot (RollRoller) tries to turn on plane via changing mass center (some minor effects have been avoided which may cause instability in robot). For circular locomotion, position of cores for producing corresponding gestures is like Figure 2.5. For simplicity of explanation it is considered that RollRoller initial form is like first gesture on beginning of left hand-side. Particularly, VT's core is located in equilibrium point and HT's core in MM part. In next Step, robot tries to maintain its forward movement despite the locomotion of horizontal tube core to boundary region on the right of two MM intersection. This cycle of movement results a circle turn with R radius. The magnitude of R is proportional to forward locomotion motion speed, horizontal tube core location and other factors. For example, we can have more sharp turns with locating the core in MM $\beta$ of horizontal tube or either clear locomotion with constant R under special circumstances.
\subsection{Angular Locomotion}
\begin{figure}[ht!]
\centering
\includegraphics[width=100 mm]{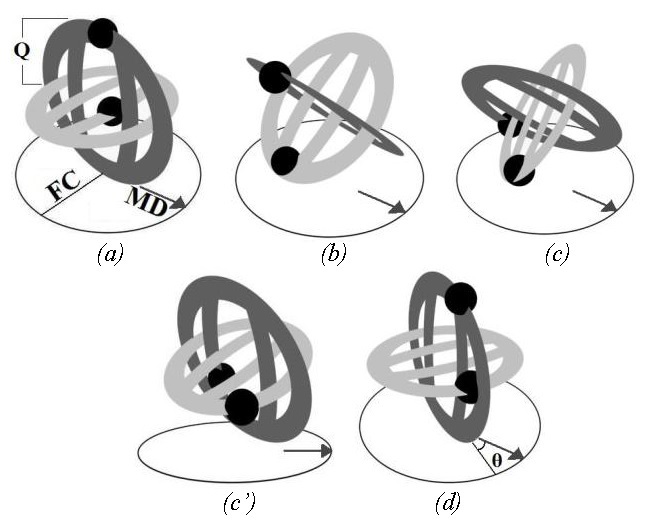}
\caption{Required gestures for angular motion.}
\end{figure}
This movement is a strong advantage of this robot. By this movement,we are able locate robot in accurate directions via center of mass diversity with minimum error. Also, there is no requirement for existence of forward direction motion like in circular locomotion we had. In contrast, other previous SMRs mostly had to have turning normally such as previous section. The proposed gestures are demonstrated in Figure 2.6. As result of having spherical shell, the robot required to operate in area dependent to radius of tubes, it is showed as fire circle (FC). Main direction (MD) defined as steady posture. Case a is initial gesture of SMR for beginning of motion. Turner (Horizontal) tube core moves to right side of robot and rotates it $45^o$ along itself in case b. Then Forwarder (Vertical) tube has $45^o$ change along Turner tube and goes backward. Eventually, for returning the robot to its main gesture, Forwarder and Turner tube's cores moves simultaneously as case c. Case c’ is been demonstrated to give better perspective about the motion. In same time, VT's core pass through the GB path. However, the core in Turner (horizontal) tube moves to momentum maker path. At the end, sphere will have $\theta$ angular change due to main direction. In here Q gives the minimal angular change respect to radius of VT in sphere but in reality this is much smaller ($22.5^o > \theta$)

\subsection{Slide Locomotion}
\begin{figure}[ht!]
\centering
\includegraphics[width=85 mm]{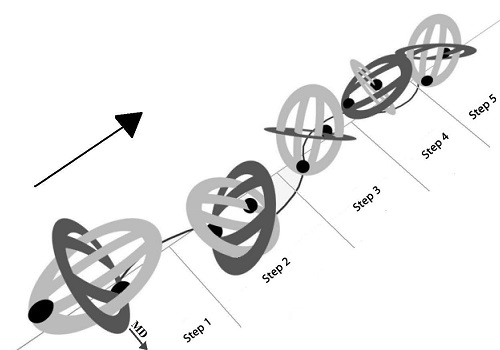}
\space \space \space (a)
\includegraphics[width=90 mm]{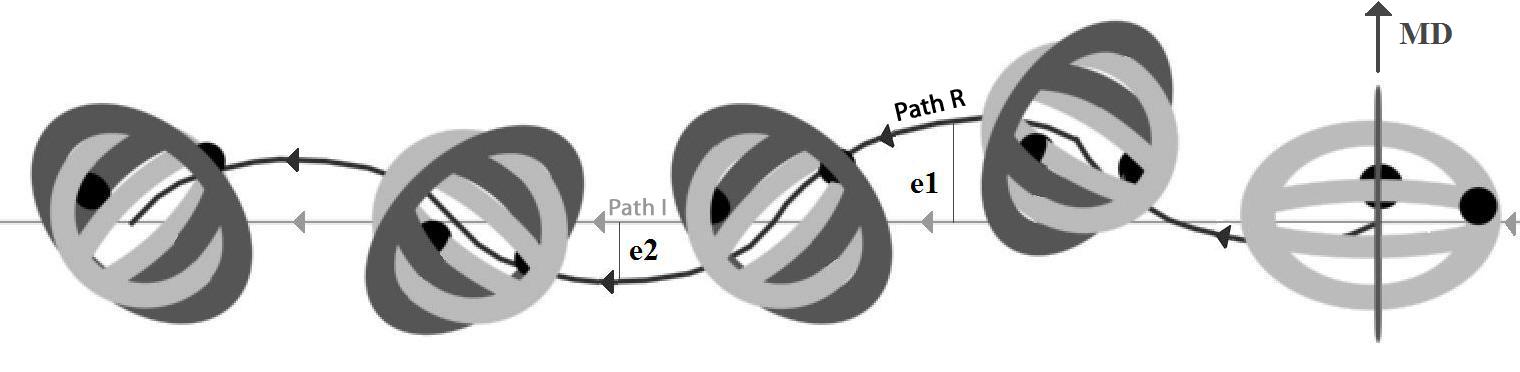}
(b)
\caption{Cycle of slide locomotion: (a).Perspective view , (b.) Top view.}
\end{figure}

This movement shows that RollRoller can move in slide direction without any required turning. Also, this movement takes it corresponding movement force with gravity change hence the possibility of accidental error considerably low. Figure 2.7 case a shows steps about manner. The algorithm of movement is as follow:

\textbf{Step 1}:	 Initializing the condition of robot.

\textbf{Step 2}: Movement of Turner Tube core by passing through right Gravity Breaker (GB) and entering the right Momentum Maker (MM) part.

\textbf{Step 3}:	Reaching the core to end of Turner Tube (TT).

\textbf{Step 4}:	Forwarder Tube (FT) core moves go through left Momentum Maker (MM) path and gives the robot displacement.

\textbf{Step 5}: Return to Step 3.

And case (b) shows the movement in clear top view manner. Path I demonstrate as ideal movement and Path R is our real posture line which ‘O’ stands for offset of sphere due to its shape.	

\subsection{Jump Locomotion}
This motion demonstrates the other unique specification of RollRoller as 4th DOF. In this motion we considered the first gesture of robot like previous ones. Then for initializing the jumping activity, the turner tube core moves to corner of tube (Equilibrium point) for giving the corresponded gesture as Figure 2.8 (a) in same time Forwarder Tube (FT) core goes through momentum maker path. Figure 2.8 (b) shows these steps.
\begin{figure}[ht!]
\centering
\includegraphics[width=50 mm]{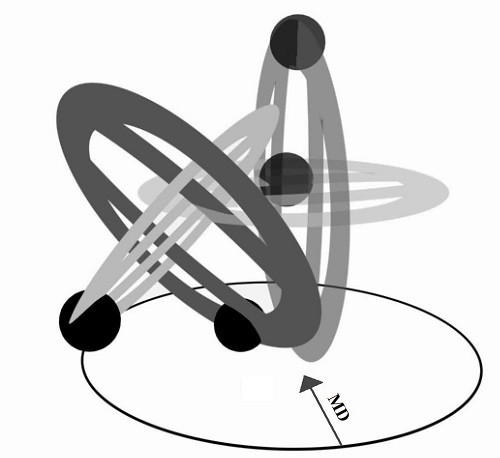}
\space \space \space (a)
\includegraphics[width=50 mm]{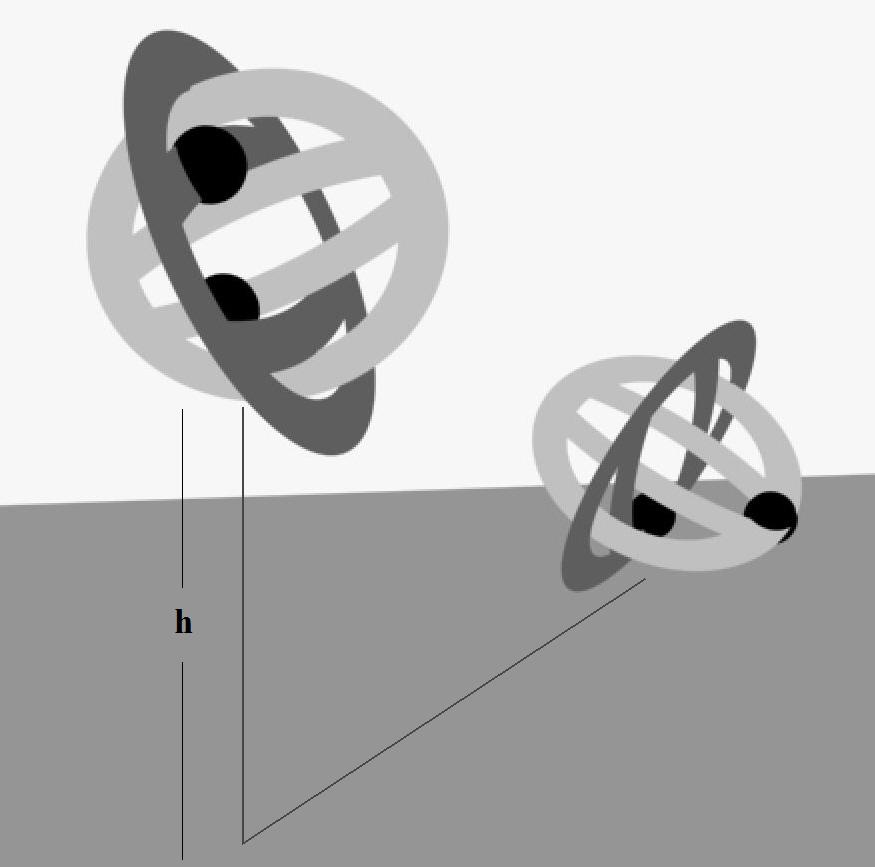}
(b)
\caption{Cycle of Angular motion (a).Initial formation. (b) Jumping Operation.}
\end{figure}

Sphere after initializing the structure as shown in Figure 2.8 (b), cores in same time moves through Gravity Breaker part and by using the torque reaction, cores meet the closed gates. Accordingly, total force leads to parabolic jumping path in air by circular turnings. The 'h' parameter demonstrates the possible height. For giving higher jump we could add other extra movements like moving cores in momentum maker paths for increasing the speed of them and then let the event take place.
\section{Electrical Sensors Overview}
We make RollRoller as semi-intelligent robot has sensation about its current position instructions. Occasional sensors are used for accurate displacement of robot. System electrical controlling diagram is demonstrated in figure 2.9. it uses three efficient sensing material (i.e. Circular sensors, Geometric positioning sensor and gravity sensor).Geometric positioning (IMU) sensor and gravity sensor are electrical circuits, whereas Circular sensors is novel sensor specialized for only RollRoller. Main controller IC collect the coming data from sensors for visualizing the current place of core, current gravity magnitude and the position of robot in 3D plane respect to base. Next, the motion control data is sent to robot. Pre-defined instruction in main controller IC, decides the steps. Lastly, motors (pressurized air producer) and gates control works proportional to programmed algorithm.	Circular Sensors are semi-physical sensors located around the tubes for giving the location of relating cores in RollRoller.

\begin{figure}[ht!]
\centering
\includegraphics[width=100 mm]{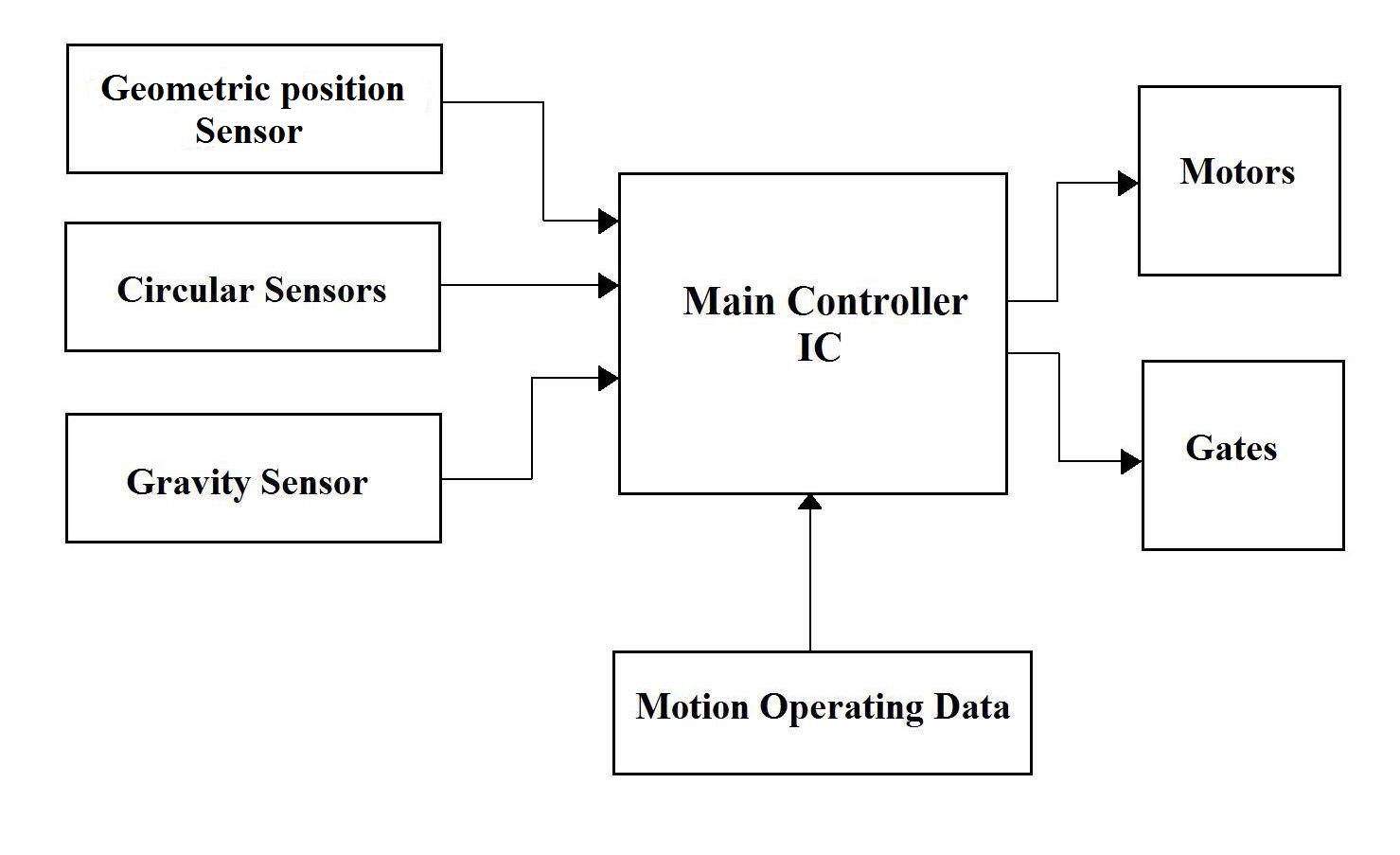}
\caption{Diagram of electrical behavior of RollRoller.}
\end{figure}

\textbf{Circular Sensor:} Circular sensors are semi-physical sensors located around the tubes for giving the location of related cores in robot. This feature is two parts circular copper as Fig.2.10, one side polarized with $V_{cc}$ Voltage and other part connected to IC port in link with one diode and resistor. Diode has put for preventing of any reverse current from IC port. While copper core passes through this instrument, changing of voltage from 0 to 1 in corresponded port locate the place of core.	
\begin{figure}[ht!]
\centering
\includegraphics[width=100 mm]{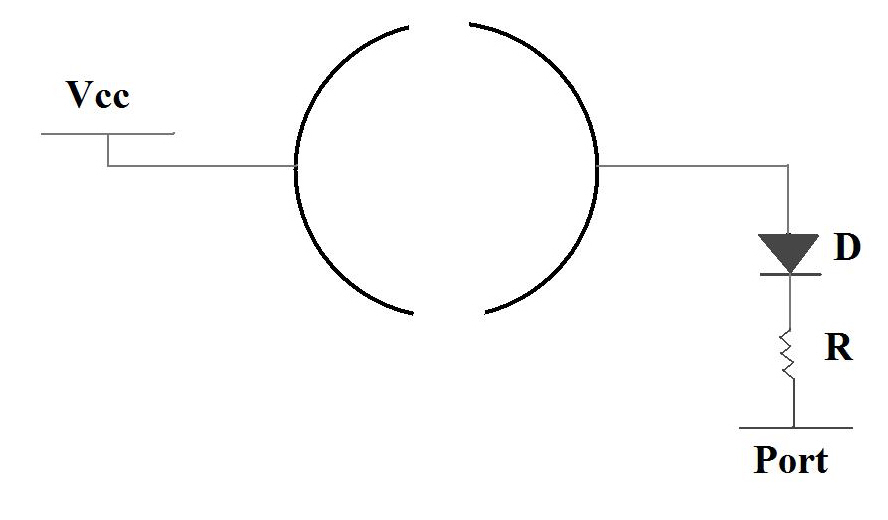}
\caption{Circular sensor.}
\end{figure}

\chapter{Mathematical Analysis}
\section{Basic Motion Dynamics Calculations}

In this chapter because "RollRoller" requires serious time and effort to calculate the practical algorithmic dynamics , only non-algorithmic ordinary dynamics [48] have been derived, the core just start moving through fixed circular tube to produce motion via gravitational and angular momentum force. The main reason behind keeping work as simple as possible is as result of nonholonomic and non-linear structure of spherical robot that it still is hard and sophisticated way to be derived \cite{nonholonomic} for general dynamical formation. As consequence, in this section, by using Euler-Lagrange method dynamics are found for containing novel model in which it will be seen that still the non-linearity of system is involved in almost all the variables as coefficient. As seen from the cycle of locomotion at previous chapter, main produced momentum is just happened in Fig.2.3 case b hence the geometric locomotion is as Fig.2.3 (the effect of case d is neglected initially).Also, the bellow assumptions are considered during :

\textbf{I}. The ground imagined without any slippery.

\textbf{II}. Homologous components have been used.

\textbf{III}. The shell outside sphere is rigid.

\textbf{VI}. The Turner Tube's core placed in equilibrium point with balancing the forward tube in straight direction.

\textbf{V}. The Forwarder Tube's core positioned on top of tube when the sphere is in statical equilibrium point.

Reference frames are denoted in which $O_0-X_0Y_0Z_0$  represents the inertial fixed reference. The moving frame connected to the center of sphere is $O_1-X_1Y_1Z_1$, which translate only with respect to inertial fixed reference. Also, $O_2-X_2Y_2Z_2$ is another frame attached to center like previous one but it just rotate only with respect to the $O_1-X_1Y_1Z_1$. (figure 3.1 shows the coordination of each unit vector with axis)
\begin{figure}[ht!]
\centering
\includegraphics[width=85 mm]{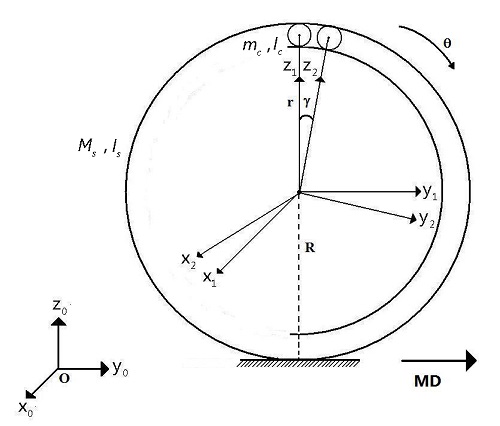}
\caption{Rolling motion geometric model along O-y.}
\end{figure}
\\

\begin{table}[htb]
\begin{center}
\begin{tabular}{cccc}
\hline 
{\footnotesize{}$\theta$} & {\footnotesize{}Rolling angle of RollRoller around x axis}\tabularnewline
{\footnotesize{}$\gamma$} & {\footnotesize{}Locomotion angle of the core}\tabularnewline
{\footnotesize{}g} & {\footnotesize{}Gravitational acceleration}\tabularnewline
{\footnotesize{}R} & {\footnotesize{}Sphere radius}\tabularnewline
{\footnotesize{}r} & {\footnotesize{}Distance between the center of the RollRoller and
the center of the core}\tabularnewline
\hline 
\end{tabular}
\end{center}
\caption{NOMENCLATURE OF DYNAMIC ANALYSIS}
\label{tab:1}
\end{table}

$D_{po}$  represents the position vector in the robot.  Angular and linear velocity vectors are $\omega_s$ and $V_s $ , also angular velocity $\omega_c$ and   linear velocity $V_c $  are relate to core.
\begin{equation}
D_{po}=rsin(\gamma+\theta)j+rcos(\gamma+\theta)k
\end{equation}
\begin{equation}
\omega_s=\dot{\theta}i
\end{equation}
\begin{equation}
V_s=R\dot{\theta}j
\end{equation}
\begin{equation}
\omega_c=(\dot{\gamma}+\dot{\theta})i
\end{equation}
\begin{equation}
V_c=(R\dot{\theta}+(\dot{\gamma}+\dot{\theta})rcos(\gamma+\theta))j-((\dot{\gamma}+\dot{\theta})rsin(\gamma+\theta))k
\end{equation}

\begin{figure}[ht!]
	\centering
	\includegraphics[width=120 mm]{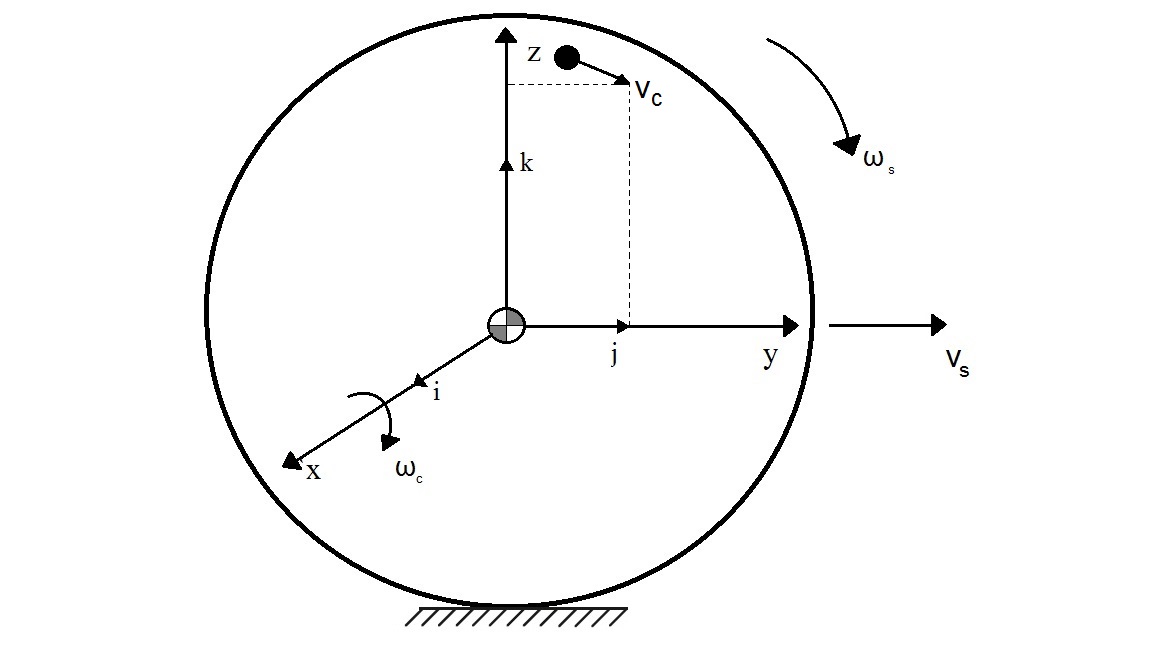}
	\caption{The unit vectors, angular and linear velocities.}
\end{figure}

The relation between sphere and core mass is as below:
\begin{equation}
m_c=\dfrac{1}{3}M_s
\end{equation}
where $M_s$  and $m_c$   are the masses of the sphere and core. This property is for emphasizing the gravity force in displacement.  For reaching the dynamic model of robot, the Lagrangian function L containing only the terms due to rotation along the O-y axis is written as follows: 
\begin{equation}
L=E_k-E_p
\end{equation}
The kinetic energy are mostly produced  angular and linear velocities from sphere and core displacement , angular momentum driven method is satisfied by preserving these energies. Gravitation energy in core is the related to potential energy ($E_p=-m_cgd_{c-z}$). In potential energy equation, the remaining distance after substituting the total vertical axis (z) from the core position gives real potential energy. 
\begin{equation}
L=\dfrac{1}{2}M_s|V_s|^2+\dfrac{1}{2}I_s|\omega_s|^2+\dfrac{1}{2}m_c|V_c|^2+\dfrac{1}{2}I_c|\omega|^2+m_cgd_{c-z}
\end{equation}
Particularly, for each of sphere and core, there are two main partition:

\textbf{1-} Created energy from object with mass (M) while containing linear velocity. ($\dfrac{1}{2}MV^2$)

\textbf{2-} Created angular energy from rotational velocity of object with its own moment of inertia (I).($\dfrac{1}{2}I\omega_s^2$)

After putting the defined equations 3.1-4 to equation 3.8, the equation 3.9 results.
$L=\dfrac{1}{2}R^2\dot \theta^2M_{s}+\dfrac{1}{2}I_s\dot \theta^2+\dfrac{1}{2}I_c(\dot\gamma+\dot \theta)^2+\dfrac{1}{2}m_c[(R\dot \theta+(\dot \gamma+\dot\theta)rcos(\gamma+\theta))^2$
\begin{equation} 
+(\dot \gamma+\dot\theta)rsin(\gamma + \theta)^2]+m_cgd_{c-z}
\end{equation}
\\In last step the equation is regulated for next stages (factorizing similar higher order variables):
\\$L=\dot\theta^2[\dfrac{1}{2}M_sR^2+\dfrac{1}{2}I_s+\dfrac{1}{2}m_cR^2]+(\dot\theta+\dot\gamma)^2[\dfrac{1}{2}I_c+\dfrac{1}{2}m_cr^2cos^2(\theta+\gamma)+$
 \begin{equation}
\dfrac{1}{2}m_cr^2sin^2(\theta+\gamma)]+\dfrac{1}{2}m_c(\dot\theta\dot\gamma+\dot{\theta^2})Rrcos(\theta+\gamma)+m_cgrcos(\theta+\gamma)
\end{equation}
\space  \\$=\dfrac{1}{2}\dot\theta^2[M_sR^2+I_s+m_cR^2]+\dfrac{1}{2}(\dot\theta+\dot\gamma)^2[I_c+m_cr^2]$
\begin{equation}
+(\dot\theta\dot\gamma+\dot{\theta^2})m_cRrcos(\theta+\gamma)+m_cgrcos(\theta+\gamma)
\end{equation}
For calculating the dynamics via Euler-lagrange equation , bellow Lagragian equations are derived for each variable ($\gamma$ and $\theta$):
\begin{equation}\\
\dfrac{d}{dt}(\dfrac{\partial L}{\partial \dot q_i})- \dfrac{\partial L}{\partial q_i}= \tau_i
\end{equation}
I. As first step $\theta$ derivation dynamic is calculated:\\
1.$\partial L/\partial\theta=-[(\dot\theta\dot\gamma+\dot\theta^2)m_cRrsin(\theta+\gamma)+m_cgrsin(\theta+\gamma)]$\\
\space
2.$\partial L/\partial\dot\theta=\dot\theta[M_sR^2+I_s+m_cR^2]+(\dot\theta+\dot\gamma)[I_c+m_cr^2]+m_c(\dot\gamma+2\dot{\theta})Rrcos(\theta+\gamma)$\\
lastly, the given generic equation modified for the relevant variable:\\ 
$\dfrac{d}{dt}(\dfrac{\partial L}{\partial \dot \theta})- \dfrac{\partial L}{\partial \theta}= 0
\space =>
$ \\ \\
$\ddot{\theta}[M_sR^2+I_s+m_cR^2+I_c+m_cr^2]+\ddot{\gamma}[I_c+m_cr^2]+m_c(\ddot\gamma+2\ddot\theta)Rrcos(\theta+\gamma)$
\begin{equation}
-(\dot\theta+\dot\gamma)^2(m_c)Rrsin(\theta+\gamma)+m_cgrsin(\theta+\gamma)=0
\end{equation}
II.Then $\gamma$ derivation is calculated in same way:\\
1.$\partial L/\partial\gamma=-[m_c(\dot\theta\dot\gamma+\dot{\theta^2})m_cRrsin(\theta+\gamma)+m_cgrsin(\theta+\gamma)]$\\
2. $\partial L/\partial\dot\gamma=(\dot\theta+\dot\gamma)[I_c+m_cr^2]+m_c\dot\theta Rrcos(\theta+\gamma)$\\
Lastly:\\
$\dfrac{d}{dt}(\dfrac{\partial L}{\partial \dot \gamma})- \dfrac{\partial L}{\partial \gamma}= \tau
\space =>
$
\begin{equation}
\ddot{\theta}[I_c+m_cr^2+m_cRrcos(\theta+\gamma)]+\ddot{\gamma}[I_c+m_cr^2]+m_cgrsin(\theta+\gamma)=\tau
\end{equation}
Input torque ($\tau$) in equation 3.13 is zero since there is no reaction force relative to sphere ( e.g. in pendulum models this value exists with negative sign $\tau$ [48]). On the other hand, the input in equation 3.14 is series of air pulses generated by pneumatic cylinders.

"RollRoller"'s compact form of motion equation can be shown as beneath matrix form:
\begin{equation}
\left[\begin{array}{ccc}
M_{11} & M_{12} \\
M_{21} & M_{22}\\
\end{array}\right]
\left[\begin{array}{ccc}
\ddot{\theta} \\
\ddot{\gamma}\\
\end{array}\right]+\left[\begin{array}{ccc} N_{11}\\ N_{21}\\ \end{array}\right]+\left[\begin{array}{ccc} G_{11}\\ G_{21}\\ \end{array}\right]=\left[\begin{array}{ccc} 0\\ \tau \\ \end{array}\right]
\end{equation}
$M_{11}= M_sR^2+I_s+m_cR^2+I_c+m_cr^2+2m_cRrcos(\theta+\gamma)  $\\$ M_{12} = M_{21} = I_c+m_cr^2+m_cRrcos(\theta+\gamma)$ ,\space $
M_{22} = I_c+m_cr^2$\\ $N_{11} = -(\dot\theta+\dot\gamma)^2(m_c)Rrsin(\theta+\gamma)$ , \space $N_{21}=0$ , $G_{11}=G_{22}=m_cgrsin(\theta+\gamma)$. 

Although, this matrix form may not give reliable clues about the role of each component on creation of RollRoller's locomotion, the $N_{11}$ coefficient can guessed as main swinging part. In particular, the "$(\dot\theta+\dot\gamma)^2$" part gives parabolic raise and the "$sin(\theta+\gamma)$" part represents the sinusoidal swinging around this parabolic increase. For the sphere this value is zero ($N_{21}$). The other coefficients that get involved in 'M' array variables are mostly constant or have considerable small effect on system manner.  Lastly, 'G' represents gravitational factors for both dynamical equations.

To Make State-space form simulation results on Matlab, The equations are compacted by defined variables.
\\a)
\begin{equation}
\ddot{\theta}[A]+\ddot{\gamma}[B]-(\dot\theta+\dot\gamma)^2(m_c)Rrsin(\theta+\gamma)+m_cgrsin(\theta+\gamma)=0
\end{equation}
\\b)
\begin{equation}
\ddot{\theta}[B]+\ddot{\gamma}[D]+m_cgrsin(\theta+\gamma)=\tau
\end{equation}
\\Definitions: 
\\$A=M_sR^2+I_s+m_cR^2+I_c+m_cr^2+2m_cRrcos(\theta+\gamma)=a+2bcos(\theta+\gamma)$
\\$B=I_c+m_cr^2+m_cRrcos(\theta+\gamma)=D+bcos(\theta+\gamma)$
\\$C=(\dot\theta+\dot\gamma)^2$
\\$b=m_cRr$
\\$a=M_sR^2+I_s+m_cR^2+I_c+m_cr^2$
\\$D=I_c+m_cr^2$
\\$h=m_cgr$
\\$E=AD-B^2$
\\
Then definitions are substituted in equations 3.16-17 as underneath: 
\\ 
\\a')
\\
\begin{equation}
\ddot{\theta}=\frac{C}{A}m_cRrsin(\theta+\gamma)-\frac{B}{A}\ddot{\gamma}-\frac{m_cgr}{A}sin(\theta+\gamma)
\end{equation}
\\
\\b')
\begin{equation}
\ddot{\gamma}=\frac{1}{D}\tau-\frac{B}{D}\ddot{\theta}-\frac{m_cgr}{D}sin(\theta+\gamma)
\end{equation}
After substituting the b' to a' we will have : 
\\
\\a'')
\\
\\$\ddot{\theta}=\frac{C}{A}m_cRrsin(\theta+\gamma)-\frac{B}{A}[\frac{1}{D}\tau-\frac{B}{D}\ddot{\theta}-\frac{m_cgr}{D}sin(\theta+\gamma)]-\frac{m_cgr}{A}sin(\theta+\gamma)$\\
\begin{equation}
\rightarrow
\ddot{\theta}=\frac{CD}{AD-B^2}m_cRrsin(\theta+\gamma)-\frac{B}{AD-B^2}\tau+\frac{B-D}{AD-B^2}m_cgrsin(\theta+\gamma)
\end{equation}
\\
\\b'')
\begin{equation}
\ddot{\gamma}=\frac{A}{(AD-B^2)}\tau-\frac{BC}{AD-B^2}m_cRrsin(\theta+\gamma)+\frac{B-A}{AD-B^2}m_cgrsin(\theta+\gamma)
\end{equation}
\\Now by defining the state-space model, the specific formulation of equations have been created ($x_1=\theta,x_2=\dot{\theta},x_3=\gamma,x_4=\dot{\gamma}$).\\

\begin{equation}
\dot x_1=x_2
\end{equation}
\begin{equation}
\dot{x_2}=bsin(x_1+x_3)[\frac{(x_2+x_4)^2D+hcos(x_1+x_3)}{E}]-[\dfrac{D+bcos(x_1+x_3)}{E}]\tau
\end{equation}
\begin{equation}
\dot x_3=x_4
\end{equation}
\\$\dot x_4=-sin(x_1+x_3)[\dfrac{((b(x_2+x_4)^2(D+bcos(x_1+x_3)))+((a-D)+bcos(x_1+x_3))h)}{E}]$
\begin{equation}
+[\dfrac{a+2bcos(x_1+x_3)}{E}]\tau
\end{equation}
As mentioned before, it is obvious even though we find out the homogeneous and basic dynamic of SMR (without using any Euler-Jacobi-Lie theorem \cite{nonholonomic}), the equations are taking the form of nonlinear (e.g. square functions and sinusoidal sums, multiplications and division).
\section{Fractional Motion Dynamics}
The previous dynamic calculation were for just only to validate robots, following known trajectory to make its main motion. In this part, to make motion more realistic the viscous friction between core, sphere and ground has been involved. The energy dissipation function is relative to damping constant and system's velocity. This will help us to understand robot better how reacts to initial steps. The definition of equation is as:
\begin{equation}
P=\frac{1}{2}\zeta(\dot{\theta}+\dot{\gamma})
\end{equation}
Hence the Lagrange-Euler equation will be updated as bellow formula:
\begin{equation}
\dfrac{d}{dt}(\dfrac{\partial L}{\partial \dot q_i})- \dfrac{\partial L}{\partial q_i}+\dfrac{\partial P}{\partial \dot{q}_i}= \tau_i
\end{equation}
Next, the dynamical equations are updated ($\dfrac{\partial P}{\partial \dot{\theta}_i}=\zeta\dot{\theta}$ and $\dfrac{\partial P}{\partial \dot{\gamma}_i}=\zeta \dot{\gamma}$):

a)\\
$\ddot{\theta}[M_sR^2+I_s+m_cR^2+I_c+m_cr^2]+\ddot{\gamma}[I_c+m_cr^2]+m_c(\ddot\gamma+2\ddot\theta)Rrcos(\theta+\gamma)$
\begin{equation}
-(\dot\theta+\dot\gamma)^2(m_c)Rrsin(\theta+\gamma)+m_cgrsin(\theta+\gamma)+\zeta\dot{\theta}=0
\end{equation}

b)
\\
\begin{equation}
\ddot{\theta}[I_c+m_cr^2+m_cRrcos(\theta+\gamma)]+\ddot{\gamma}[I_c+m_cr^2]+m_cgrsin(\theta+\gamma)+\zeta \dot{\gamma}=\tau
\end{equation}
And matrix compact form is been changed.
\begin{equation}
\left[\begin{array}{ccc}
M_{11} & M_{12} \\
M_{21} & M_{22}\\
\end{array}\right]
\left[\begin{array}{ccc}
\ddot{\theta} \\
\ddot{\gamma}\\
\end{array}\right]+\left[\begin{array}{ccc} N_{11}\\ N_{21}\\ \end{array}\right]+\left[\begin{array}{ccc} G_{11}\\ G_{21}\\ \end{array}\right]=\left[\begin{array}{ccc} 0\\ \tau \\ \end{array}\right]
\end{equation}
$M_{11}= M_sR^2+I_s+m_cR^2+I_c+m_cr^2+2m_cRrcos(\theta+\gamma)  $\\$ M_{12} = M_{21} = I_c+m_cr^2+m_cRrcos(\theta+\gamma)$ ,\space $
M_{22} = I_c+m_cr^2$\\ $N_{11} = -(\dot\theta+\dot\gamma)^2(m_c)Rrsin(\theta+\gamma)++\zeta\dot{\theta}$ , \space $N_{21}=\zeta \dot{\gamma}$ , $G_{11}=G_{22}=m_cgrsin(\theta+\gamma)$. 
Then for State-space implementation as previous calculations the extra variables are placed in equation:

\begin{equation}
\dot x_1=x_2
\end{equation}
$\dot{x_2}=bsin(x_1+x_3)[\frac{(x_2+x_4)^2D+hcos(x_1+x_3)}{E}]+[\frac{x_4(D+bcos(x_1+x_3)-x_2D}{E}]\zeta$
\begin{equation}
-[\dfrac{D+bcos(x_1+x_3)}{E}]\tau
\end{equation}
\begin{equation}
\dot x_3=x_4
\end{equation}
\\$\dot x_4=-sin(x_1+x_3)[\dfrac{((b(x_2+x_4)^2(D+bcos(x_1+x_3)))+((a-D)+bcos(x_1+x_3))h)}{E}]$
\begin{equation}
[\frac{x_2(D+bcos(x_1+x_3)-x_4(a+2bcos(x_1+x_3))}{E}]\zeta+[\dfrac{a+2bcos(x_1+x_3)}{E}]\tau
\end{equation}
From the latest form of state-space we can see , there is huge influence of friction between objects. As consequence, during the design of RollRoller VT , HT and cores, choosing the type of materials like Nylon 12 and copper alloy decrease the likelihood of failure in robots motions since as the friction between these objects increase (will seen on results of Matlab Simulation) input consistency decreases faster and this means lots of energy losses. 
\chapter{Visualization Results And Simulink}
\section{Matlab Simulink}
\subsection{Basic Motion simulation}
To obtain the SMR's basic forward direction simulation , table 4.1 parameters are designed for both non-fictional and frictional terrain. The states' initial conditions are considered as $X=[x_1$ \space $x_2$ \space $x_3$ \space $x_4]=[0$ \space $ 0 $ \space $0 $ \space $ 0]$.The values for core's moment of inertia ($I_c$) and sphere's moment of inertia ($I_s$) have been calculated from below following equations [50] and table:
\begin{equation}
I_s=\frac{2}{3}M_sR^2
\end{equation}
\begin{equation}
I_c=\frac{2}{5}m_cr^2
\end{equation}

\begin{table}[htb]
\begin{center}
\begin{tabular}{cccc}\hline
Variable & Quantity & Variable & Quantity\tabularnewline
\hline 
g & 9.8 $m/s^2$ & R & 0.360 $m$\tabularnewline
$M_s$ & 3 $kg$ & r & 0.317 $m$\tabularnewline
$m_c$ & 1 $kg$ & $I_c$ & 0.0402 \tabularnewline
$\tau$ & 1 $kg.m/s^2$& $I_s$ & 0.2592 \tabularnewline
$\zeta $ & .8  & $r_c$ & .0215 m\tabularnewline
\hline 
\end{tabular}
\end{center}
\caption{VISUALIZATION SETUP OF "RollRoller".}
\label{tab:1}
\end{table}

Next , The block diagram of two main nonlinear state equations from equation 3.23-24 are made on Matlab Simulink as figure 4.1. Also, specific parameters are defined to make calculation visor with redefining constants in equations.
\begin{figure}[ht!]
\centering
\includegraphics[width=140 mm]{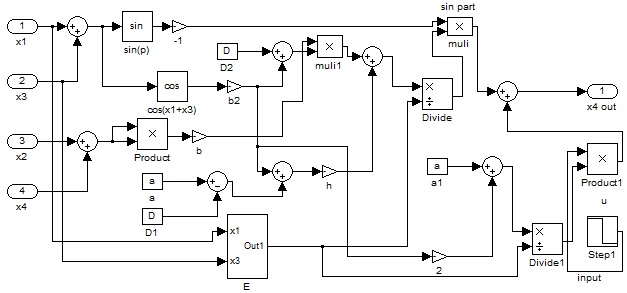}
\includegraphics[width=140 mm]{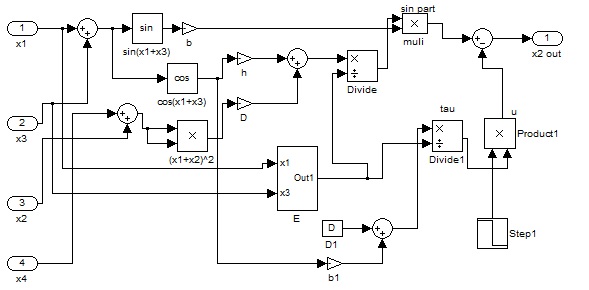}
\caption{The inner block diagram of $x_2$ and $x_3$ state-equations.}
\end{figure}
\\ Definitions: 
$
\\b=Rrmc
\\h=m_cgr
\\a=R^2 Ms+I_s+I_c+(R^2 m_c)+(r^2*m_c)
\\D=I_c+(r^2m_c)
$
\\
Particularly, the sub-block diagram for function 'E' also defined as figure 4.2.As final stage of designing, the total state-space model is created in figure 4.3.
\begin{figure}[ht!]
\centering
\includegraphics[width=100 mm]{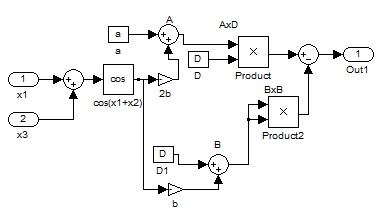}
\caption{Block diagram of function E.}
\end{figure}
\begin{figure}[ht!]
\centering
\includegraphics[width=80 mm]{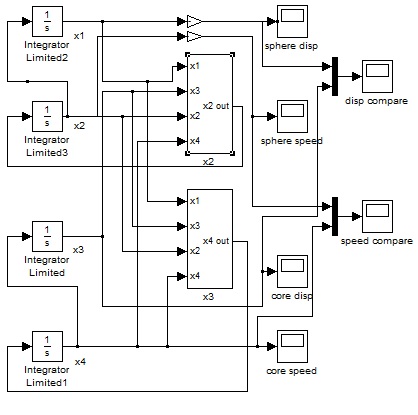}
\caption{Generic block diagram of state-space model.}
\end{figure}
\begin{figure}[ht!]
	\centering
	\includegraphics[width=140 mm]{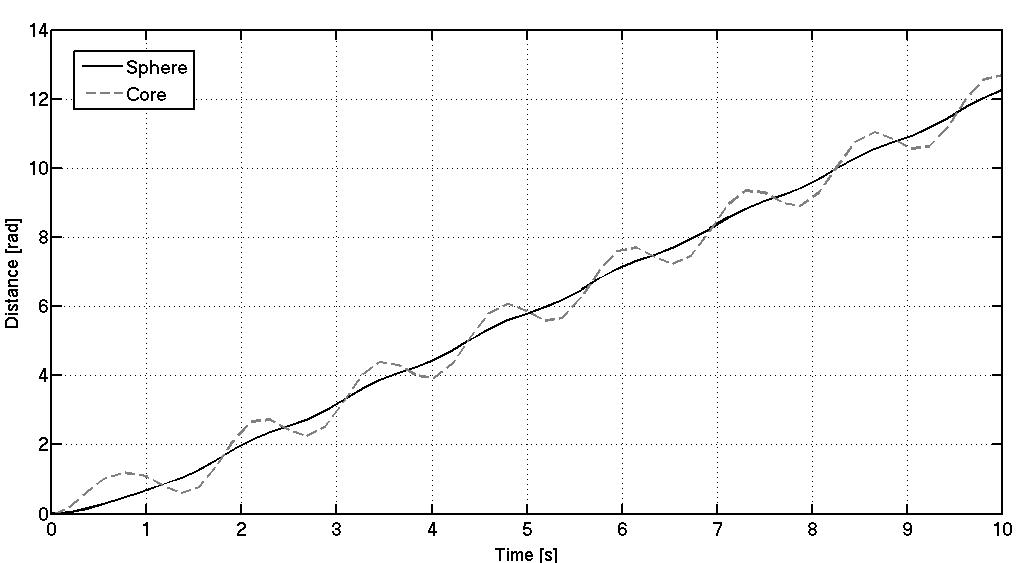} (a)
	\includegraphics[width=140 mm]{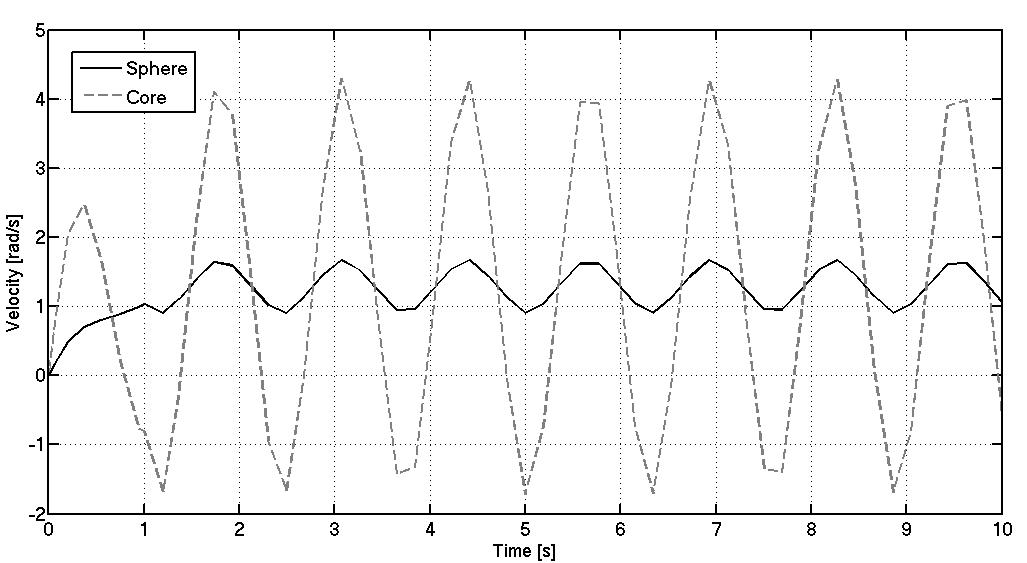} (b)
	\caption{(a). Displacement of core and sphere (b). Velocity for both core and sphere.}
\end{figure}

By using scopes from output states, displacement and angular velocity respect to radian during time (sec) for total SMR and core are obtained on Figure 4.4 ( before generating results the Matlab code in Appendix A has been run to load defined constants in block diagram). Specially in figure 4.4.a to have output the input $\tau$ only given for just 1 sec as step pulse input ( Figure 4.5) then the core starts moving through the pipe from predefined initial location as figure 2.3 a in section 2.2.1.
\begin{figure}[ht!]
	\centering
	\includegraphics[width=100 mm]{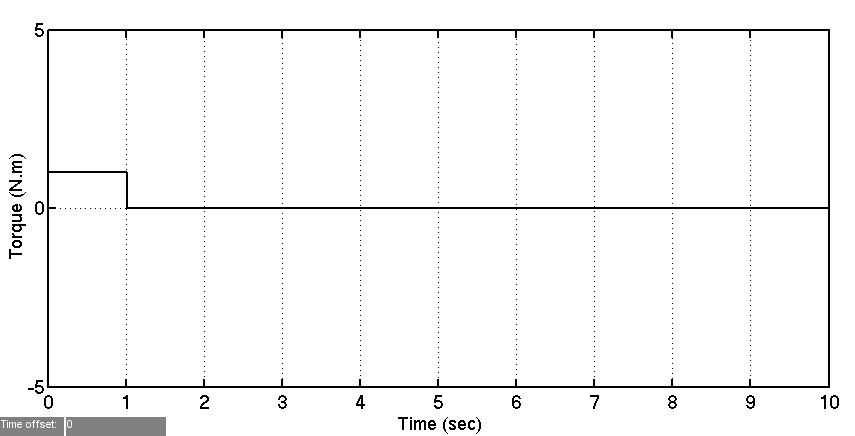}
	\caption{Pulse input to system ($\tau$).}
\end{figure}

\begin{figure}[ht!]
	\centering
	\includegraphics[width=140 mm]{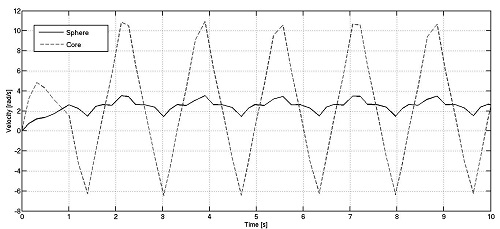} (a)
	\includegraphics[width=140 mm]{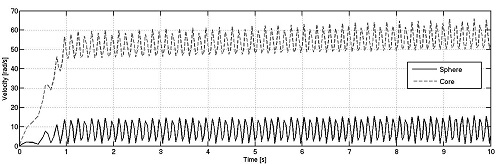} (b)
	\caption{ (a). Velocity response of sphere and core states with double pulse input to system ($2 \tau$) (b). Velocity response of sphere and core states with higher pulse input to system ($6 \tau$)}
\end{figure}

Swinging in displacement of core is reflecting the physical characteristic of the circular tube that it just stays in constant amplitude range. The total sphere via gaining force from angular momentum and gravity let the central mass be diverse during process, begins to move the sphere in direct path. Because the robot contains simple mechanism the requirement of serious control methods are avoided. RollRoller doesn't have any serious fluctuation in it's displacement ,although main algorithm including Gravity Breakers, haven't been implemented to have more smooth and damped movement. Figure 4.4.b  demonstrates velocity of core and sphere. Both velocities shifted proportional to input torque ($\tau=1 N.m$). As a point, it is understandable from results, velocity relation of core with sphere is somehow beneficial but there is sinusoidal fluctuation in velocity of sphere which will be sorted out by algorithm in section 2.2.1 as said.  Furthermore, the downhill amplitude of sphere's speed is almost 10\% of uphill. This is mathematical proof for direct path movement due to speed results. 

On the other hand, the velocity graph tells us by increasing the speed of core the total SMR just gets the small amount of that angular momentum fluctuation. To validate this prediction, the value of input torque ($\tau$) was increased to 2 N.m. The speed of core and sphere achieved acceptably as expected in figure 4.6 a. In the graph, scheme after shifting the SMR's  speed to around the 2-2.2 $rad/s$, the value of each uphill downhill is approximately same as the results from $\tau=1$ ($A_{\tau = 1}=A_{\tau = 2}=.8$ uphill and $A_{\tau = 1}=A_{\tau = 2}=.2$ downhill ). Nevertheless, without using proposed algorithm this results will look acceptable until certain level of input. When torque limitation exceed , sphere will contain waving manner ( Figure 4.6 b). Figure 4.6 b explains this result with having $ \tau = 6$ \space $ N.m $ input.  Eventually, this subsection validate the results about basic clear locomotion of sphere.

\subsection{Simulation of Motion with Fraction}

\begin{figure}[ht!]
\centering
\includegraphics[width=120 mm]{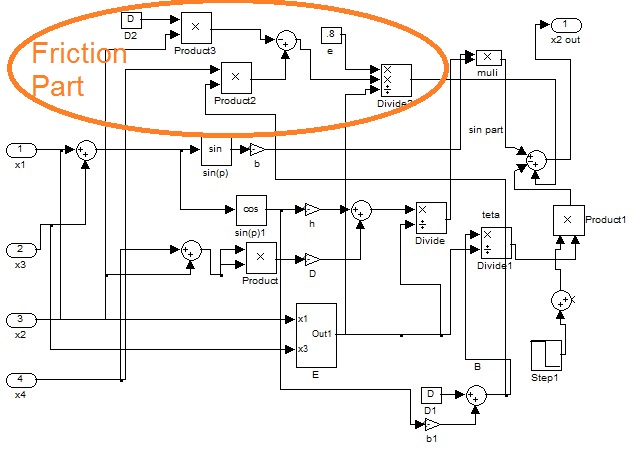} 
\includegraphics[width=120 mm]{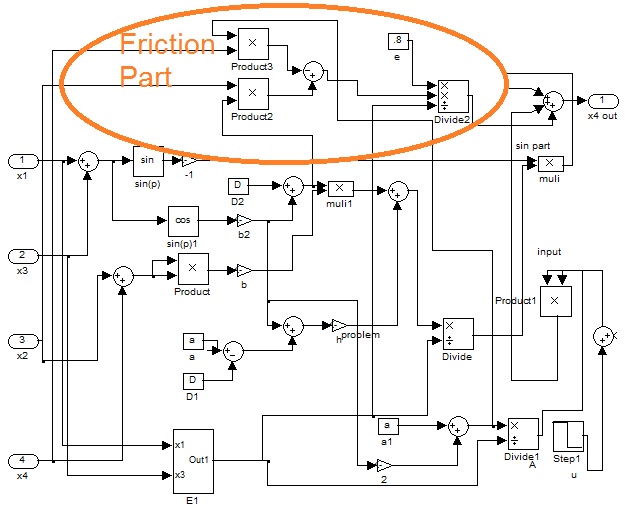} 
\caption{The inner updated block diagram of $x_2$ and $x_4$ state-space models.}
\end{figure}

A few minor changes have taken place in this simulation, for instance the $x_2$ and $x_4$ block diagrams were updated by extra variables as equations 3.32 and 3.34. The enhanced parts in $x_2$ and $x_4$ block diagrams from figure 4.1 are represented on figure 4.7 ( The orange circled parts are the updated parts to simulation). Clearly, the friction parts have influence directly to all the states,it means more the friction goes up, the convergence happens faster. As consequence, friction must be balanced in system  as reflects the main reason behind using Nylon 12-copper viscosity value. Although, the viscosity of this value is about .63, it is considered .8 to see results in worst conditional environment.
\begin{figure}[ht!]
\centering
\includegraphics[width=140 mm]{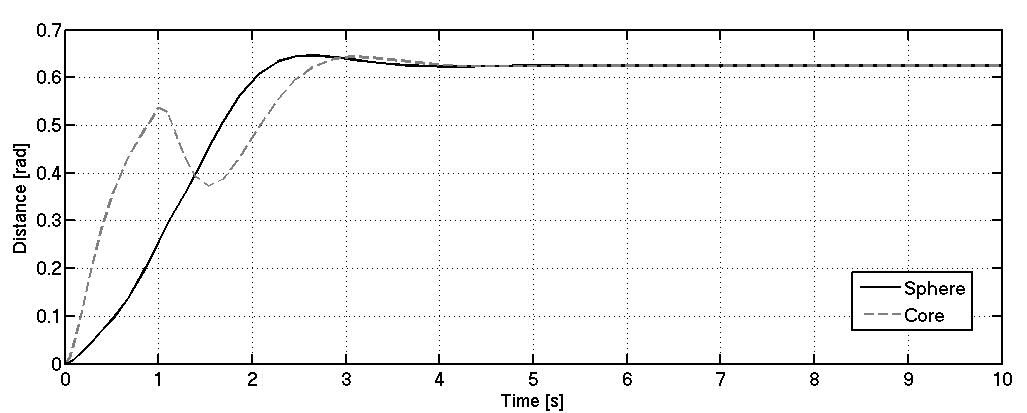}
\includegraphics[width=140 mm]{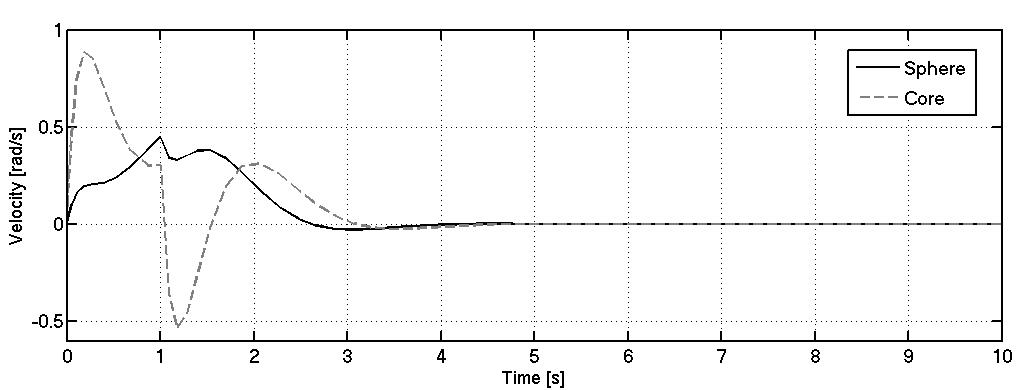}
\caption{(a). Displacement of core and sphere with friction (b). Velocity for both core and sphere with friction.}
\end{figure}

The results are illustrated on figure 4.8. The movement of sphere and core shows that the core ( dashed line) due to initial input force starts moving through tube. Next, By passing through MM (e.g. $\alpha$ as figure 2.4 ) it entered to next MM ( e.g. the opposite pipe $\beta$). Thus , because of viscous friction in surfaces the core motion damped and it stopped in middle of MM $\alpha$ tube, figure 2.3 (b) shows it. The perfect minor step displacement in sphere is demonstrated on figure 4.8 (a). By looking at the results, we can conclude that this robot can have acceptable performance without any complex control methods. In next graph 4.8 (b) the velocity of core increased as result of 1 sec pulse input, then it has changed direction corresponding to turning in tube and lastly died to zero. About the sphere the issue is a little different, although it takes the positive velocity until assistance of input (1 sec) and core motion , it has  sudden decrease ( approximately 0.1-.2) as next phenomenon. And lastly, it reaches to zero after 3 sec.

Again with having look to the results , it is obvious that the proposed morphology for our design is completely logical and implementable on practice. Following given trajectory is confirmation of true results about mathematical part.

\section{Adams/View Numerical Simulation}
\subsection{Direct Generic Motion}
To obtain the numerical simulation of "RollRoller" in forward direction motion on Adams/View , firstly the model of "RollRoller" designed on SolidWorks [51] (Figure 4.9). Particularly, in designing process the type of material (e.i. Nylon 101) and center of mass of total object are highlighted tasks having serious contribution in simulation procedure.
\begin{figure}[ht!]
\centering
\includegraphics[width=80 mm]{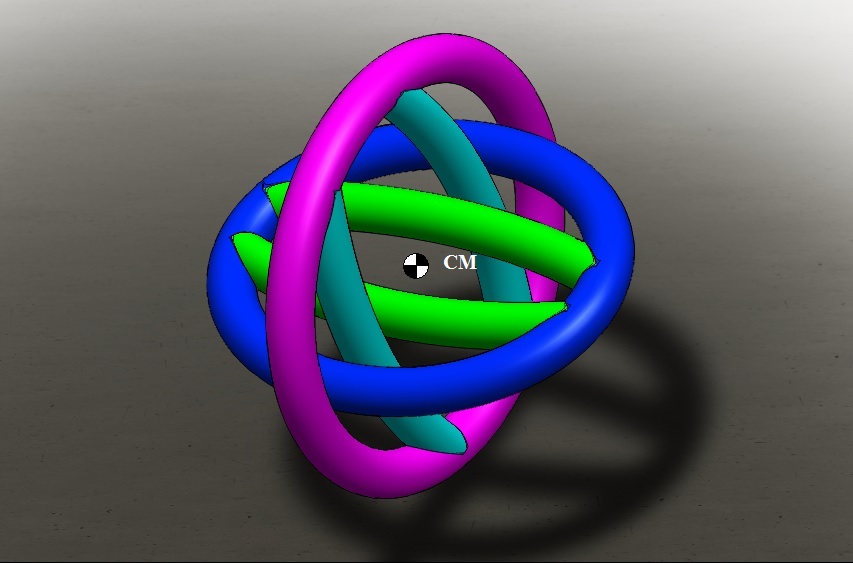}
\includegraphics[width=120 mm]{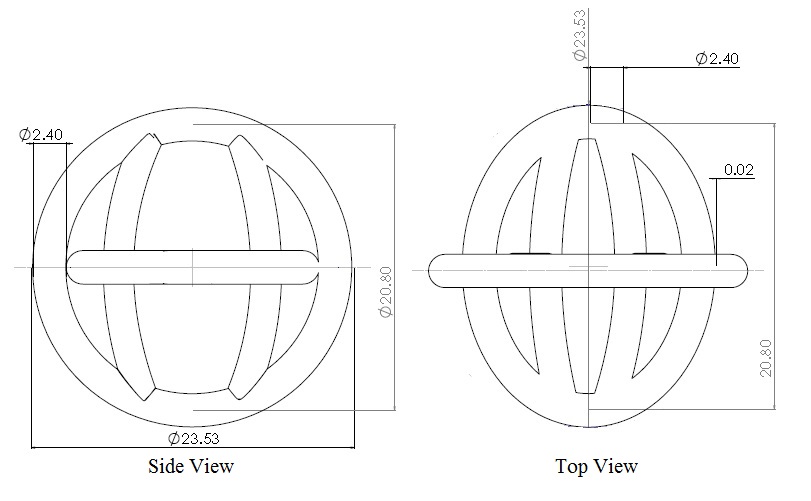}
\caption{Solidworks Design (Inches Unit).}
\end{figure}
And at end, some required parts was created on Adams/View program itself to improve the simulation, figure 4.9 shows the latest model of SMR. In this design, because giving required input air force of core to have movement inside the tube was so sophisticated and imperfect hence problem has been redirected with giving the force with another core physically. In other words, turning the pendulum-like object connected to beams from both sides of spherical shell (The yellow part in figure 4.11).Particularly, this part is lack of any weight or friction toward any main parts (e.g. Sphere, tube and Core) pushing the main core with another imaginary core. Before starting main simulation, several tasks have been followed [52]:\\
1. Plane ground was drawn with opposed y axis gravitation and locked to main ground.\\
2.The designed assembled parts were imported to Adams/view.	

2.a. HT's GB and MM pipes were locked to each other (as figure 4.10 a ). In this designing procedure, the required parts are chosen and typology of is chosen as fixed.

2.b. VT's GB and MM pipes were  locked to each other.
 
2.c. VT was locked to spherical shell.

2.d. Due to problematic simulation results in plastic density implementation the 
material in simulation considered as glass (Density = $2595 \frac{kg}{m^3} $) for HT and VT pipes (Figure 4.10 c).
	\\
3. The core and pendulum actuator were designed.

3.a. Mass of pendulum's each part considered $0.1451268544$.
	
3.b. Joint Connection was made between pendulum and  was linked beams to Sphere (Figure 4.10 b).

3.c. Torque was defined for designed pendulum as figure 4.10 d( $T=-25 N.m$ ).

3.d. The core was designed with $r= 2.85$ cm, $m_c=\frac{1}{3}M_s=15$ kg and with tungsten material ( Density = $ 1.9e+004 \frac{kg}{m^3} $ ) .

3.e. The relation between core and VT pipes was created. ( $Stiffness = 1.0e+006 \frac{N.M}{rad}$ , $Damping=1.0e+005$ and  $PD=1.0e-004$).
	\\
4. Plane,  spherical shell and other required contacts were	 created between parts. \\

\begin{figure}[ht!]
\centering
\includegraphics[width=65 mm]{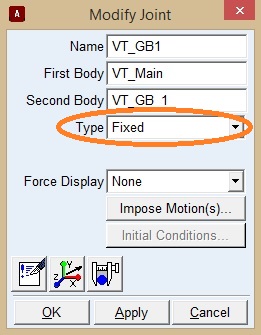}   (a)
\includegraphics[width=65 mm]{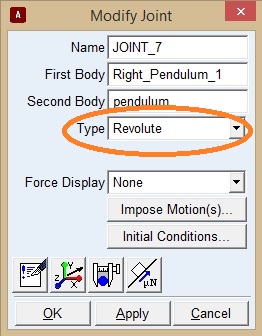}   (b)
\includegraphics[width=65 mm]{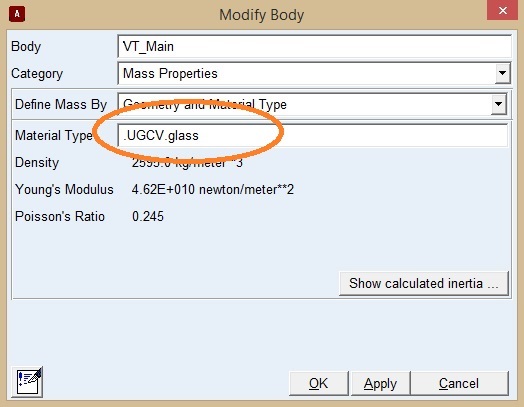}   (c)
\includegraphics[width=65 mm]{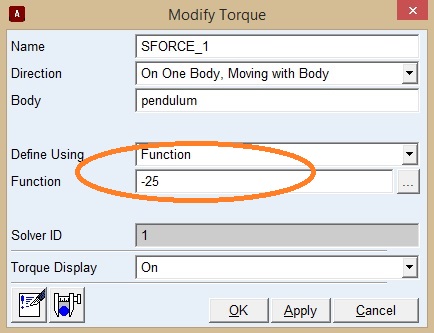}   (d)

\caption{(a). Fixed Joint between GB and VT (b). Revolute joint between Control gate and joint (c). Body properties of VT (d).Input torque to pendulum. }
\end{figure}

All created relations are demonstrated at appendix B in Figure B.1.It is considered core starting motion at equilibrium point on top of tube (i.e. case 'a' of figure 2.3). Additionally, simulation takes place within 10 seconds with sampling time of .0001. Then continuously, the motivated core follows all the given phases at figure 2.3 except case 'd' which is related to entering to GB pipes. After playing simulation and transferring it to Adams/PostProcessor, results are achieved about displacement and speed as figure 4.12. It is obvious the plotted graphs on Adams/PostProcessor are exactly tracking the calculated dynamics' outputs (only it requires scaling factors due to change in density, mass and force definitions). In conclusion, the calculated dynamics results are totally converging the visualized robots outputs.

\begin{figure}[ht!]
\centering
\includegraphics[width=140 mm]{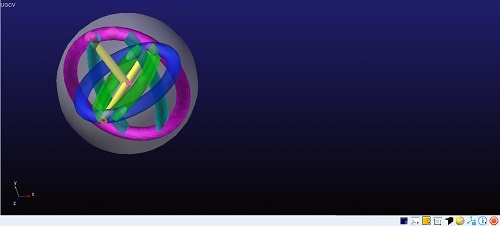}
\caption{RollRoller's Adams/view simulation design.}
\end{figure}
\begin{figure}[ht!]
\centering
\includegraphics[width=140 mm]{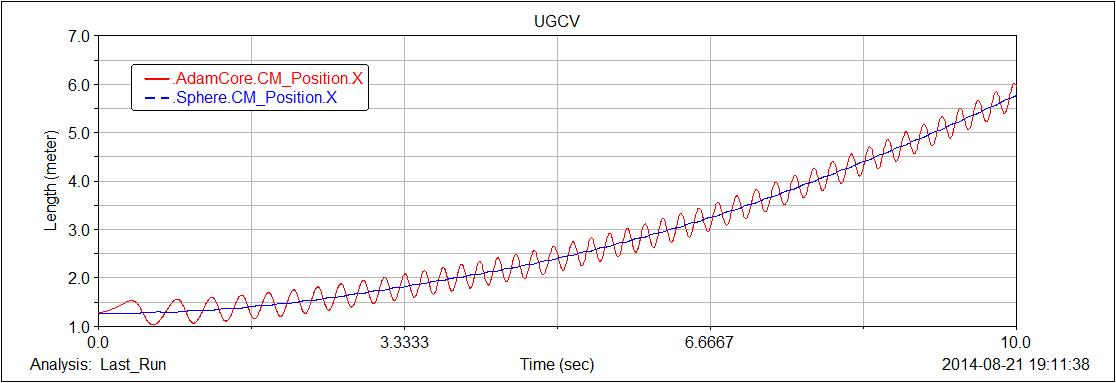} (a)
\includegraphics[width=140 mm]{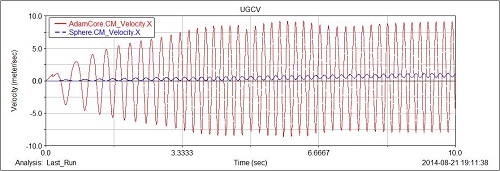} (b)
\caption{(a). Displacement of core and sphere result on Adams/PostProcessor (b). Velocity for both core and sphere result on Adams/PostProcessor.}
\end{figure}

Because the translation momentum in Y axis is one of essential parameters in giving motion to robot, the figure 4.13 illustrated about the results for sphere during this interval. As obvious from the output in this sense, although our basic motion was perfectly satisfying , the robot suffering from keeping its angular momentum in conservative mode since there is loss from jumping ups and downs of sphere (Figure 4.13). There are two main reasons for this phenomena, numerical errors that took place during simulation and imperfect motion algorithm. Nevertheless, by using main algorithm motion (figure 2.3) it is able to be sorted out because it rectifies sinusoidal behavior to more linear form, this characteristic will put under scope later on. In addition to that, this mechanism will result more rectified motion and velocity for SMR.
 (e.g. pendulum structured models).
\begin{figure}[ht!]
\centering
\includegraphics[width=140 mm]{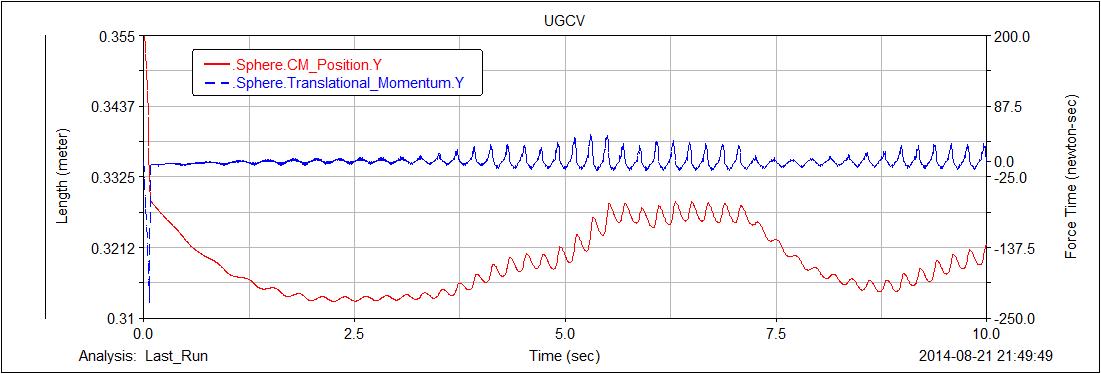}
\caption{The translational momentum and displacement in Y axis.}
\end{figure}
Another strong point of RollRoller is increasing angular acceleration as Figure 4.14. The angular acceleration around z axis demonstrating that it is able to gather speed in which for most of it's own type was impossible due to high likelihood of instability in IDU and total SMR motion hence they mostly developed in constant speed. For examples, as explained in introduction part, many driving methods with previously defined mechanisms, specially pendulum mechanism were not able to gain higher speed and they have speed and acceleration limits as center of mass is primarily located in connected weights (Figure 1.7 b and 1.8). And if actuator of IDU required to produce the higher speed during motion, it has to move its center of mass in circular path, and it will cause motor to be unable to control, while it is located in half upper surface of sphere. As result, it will cause robot to have unpredictable and unstable manners. This shows the power point of "RollRoller" novel mechanism. accordingly, proposed SMR mechanism structure can relocate the center of mass inside pipes with linear and minimum energy requirements.
\begin{figure}[ht!]
\centering
\includegraphics[width=140 mm]{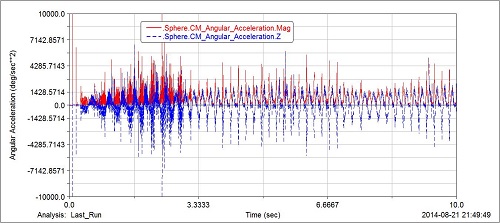}
\caption{The angular acceleration of total magnitude and Z axis.}
\end{figure}

\subsection{Algorithmic Forward Direction Motion}
In this subsection, after explaining the basic proposed novel mechanism and validating via previous simulation results. Time comes to modify the RollRoller's own motion results by letting robot follows the main created algorithm on section 2.2.1. To create required numerical simulation bellow deeds are added to previously existed model.

\textbf{I)} Gates are designed and placed as figure 4.15. Designing real implementable gate controllers in figure 2.2.(c')  were time consuming process so the more simplified one with same functionality is created. In this model, Locks has duty to stop the gate in required locations. And there is joint connecting the gate with sphere's body. Furthermore, the gate functions in either II or III that let core pass from one path only. In this simulation due to previous analysis in forward motion(reverse has same structural manner), only two gates controller are used ( mass of gate , joint and locks are considered small $M_{GC}=1.0e-009$ not to effect motion). 
\begin{figure}[ht!]
\centering
\includegraphics[width=73 mm]{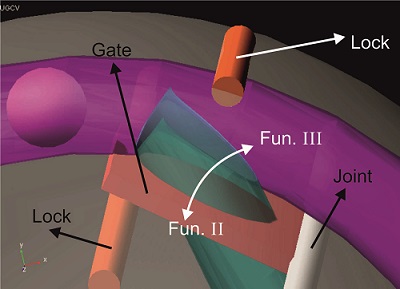} a)
\includegraphics[width=63 mm]{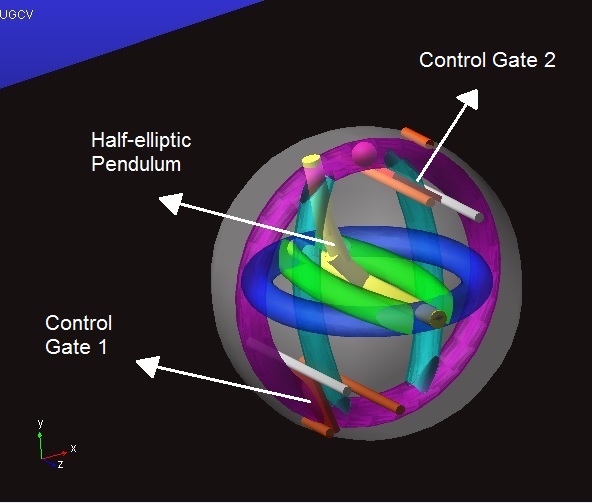} b)
\caption{(a) The special Controller gate for Adam's simulation (b).Modified simulation model.}
\end{figure}
Also, the functioning code of each gates has been written as underneath:

$FirstGateFunction=-2*(STEP( Time , 0.001 , 0.95 , .002 , 1)$
\begin{equation}
-(2*STEP( Time , 1.55 , 0 , 1.55001 , 1)))
\end{equation}

$SecondGateFuncton=2*(STEP( Time , 0.001 , 0.95 , .002 , 1)$
\begin{equation}
-(2*STEP( Time , 1.3 , 0 , 1.3001 , 1)))
\end{equation}

In this operator equation, the standard step  parameters are $STEP( x , x_0 , h_0 , x_1 , h_1 )$. Particularly, the x as time variable for this coding is defined. Consequently, $x_0$ and $x_1$
are the initial step time of first value ($h_0$) and last value ($h_1$). The paramters in each function has been found by doing the simulation continuously, this phenomenon was the reason due to not having any sensor and controller connection to "RollRoller" to have location realization. The reference step configurations are shown on figure 4.16. 
\begin{figure}[ht!]
\centering
\includegraphics[width=67 mm]{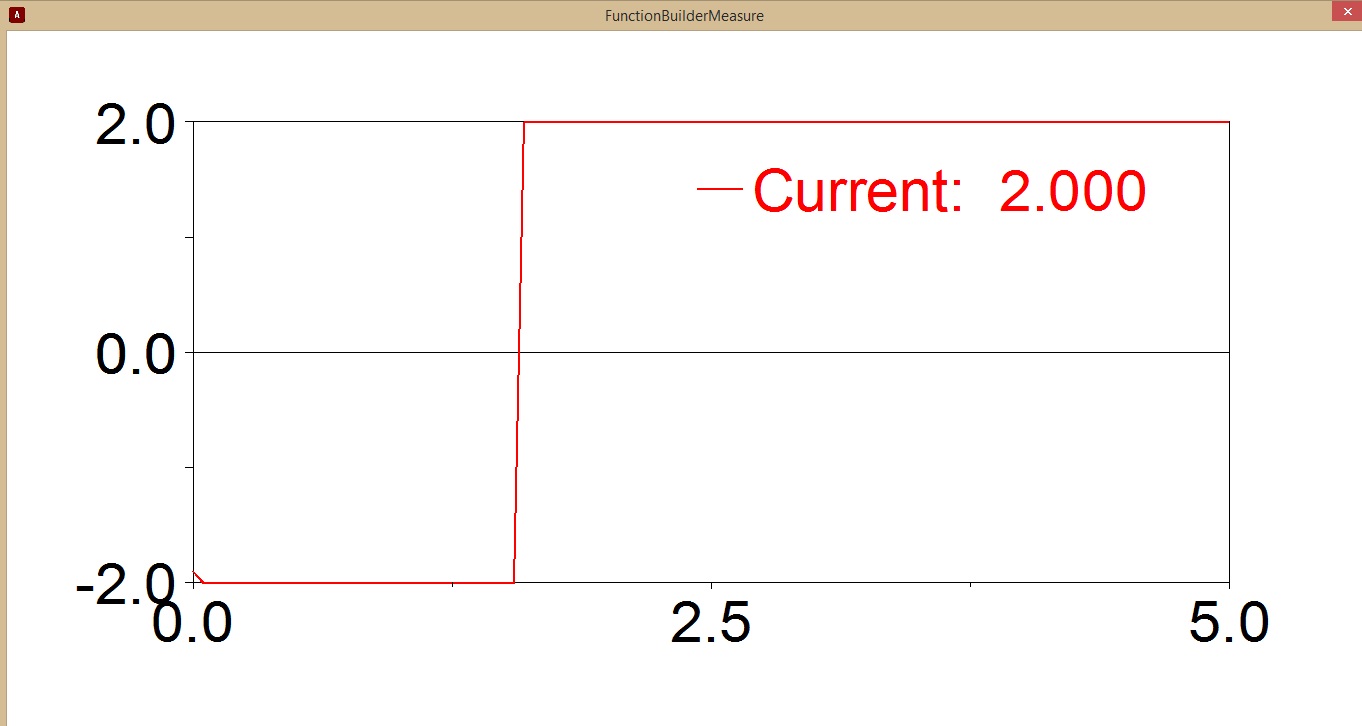} a)
\includegraphics[width=68 mm]{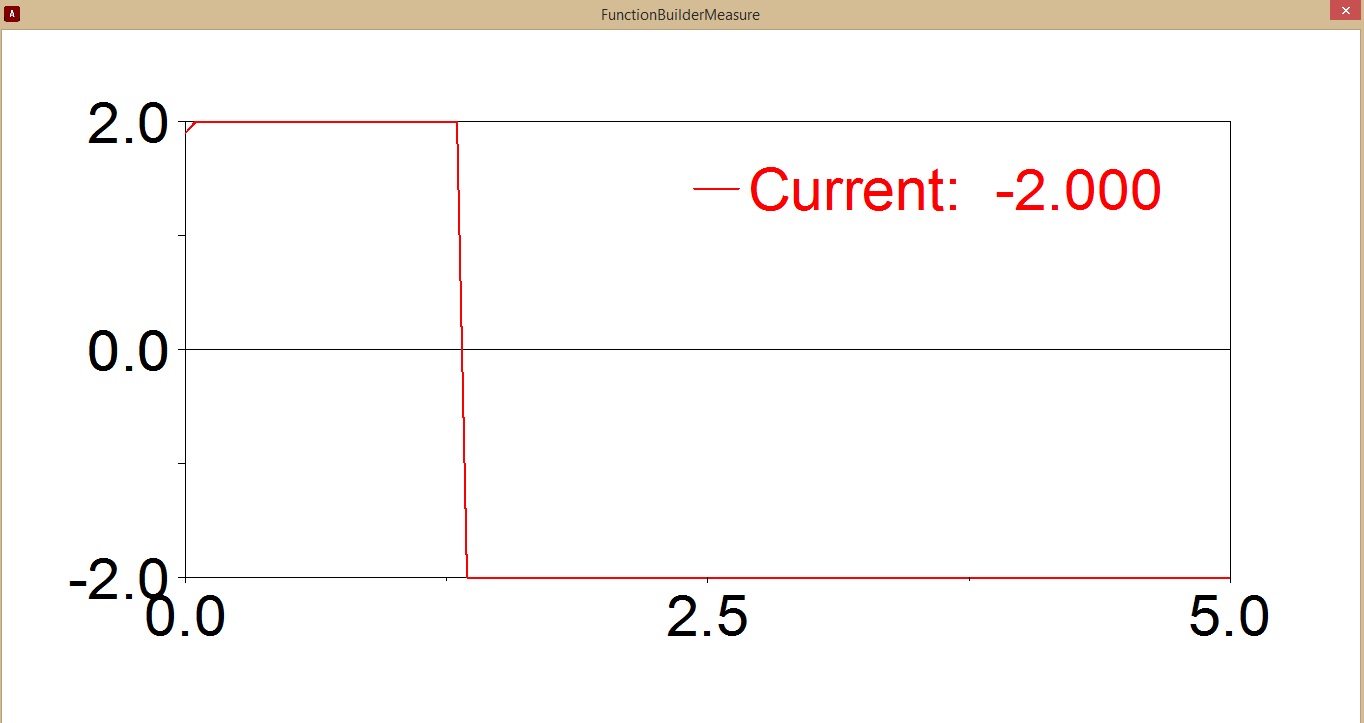} b)
\caption{(a). Control Gate 1 reference input (b). Control Gate 2 reference input.}
\end{figure}
Addition to this, the saturation of outer surface locks, located on sphere doesn't affect simulation since there is no relativity between ground and locks connectors (figure 15.a).

\textbf{II)} The pusher pendulum replaced with elliptic-like formation to prevent stock/Jamming during simulation between core, pendulum and VT.

\textbf{III)} The input torque is increased to $38.5 N.m$. Because after playing the simulation it is observed, change in form of pendulum decrease its effectiveness to push the core and it is not able to push the core in some cases (GB pipes).\\
All of the extra added bodies, connectors and motivation variables are demonstrated in Appendix B figure B.2. Figure B.3 also shows all the relativities of parts with each other in more graphical way.

\begin{figure}[ht!]
\centering
\includegraphics[width=67 mm]{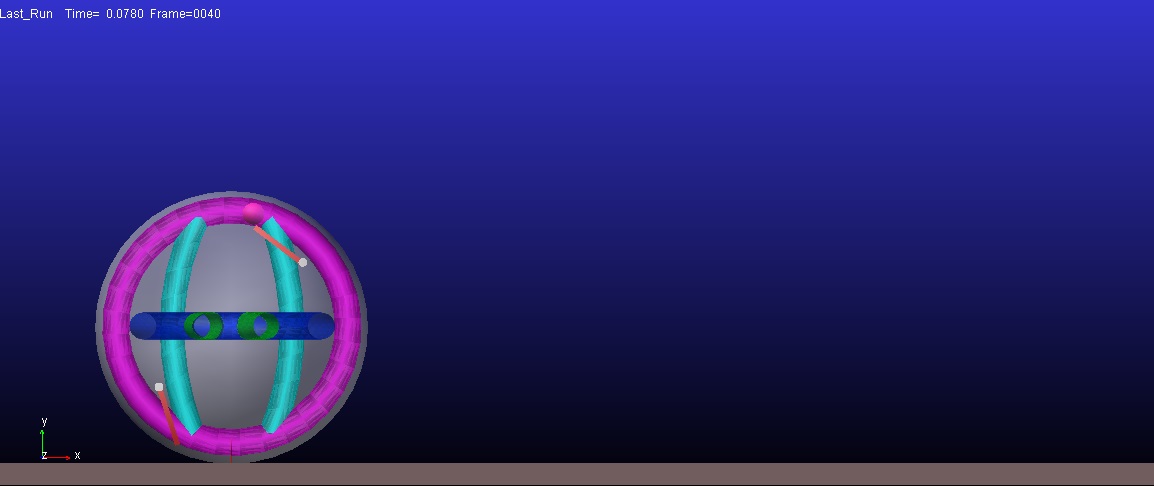} 01
\includegraphics[width=67 mm]{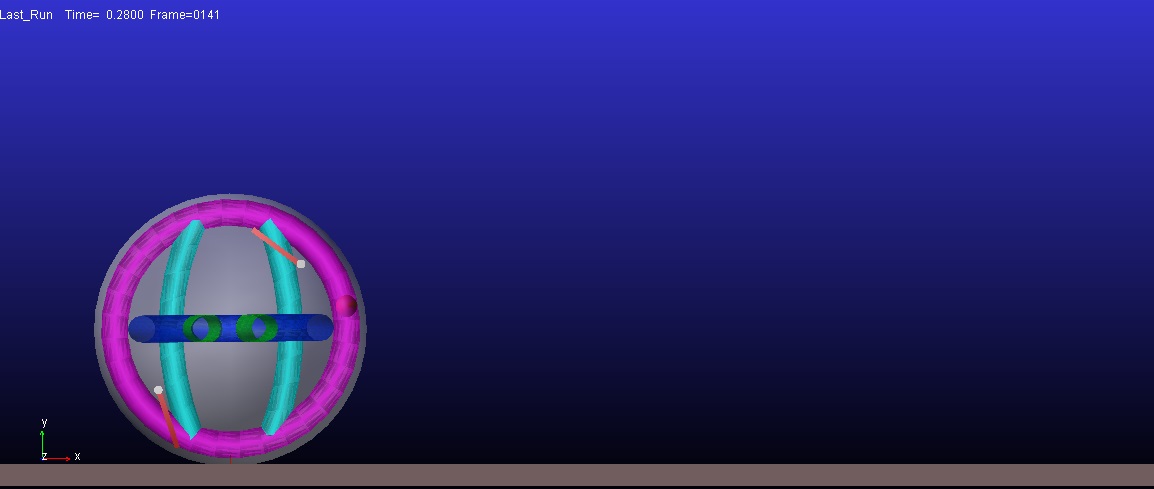} 02
\includegraphics[width=67 mm]{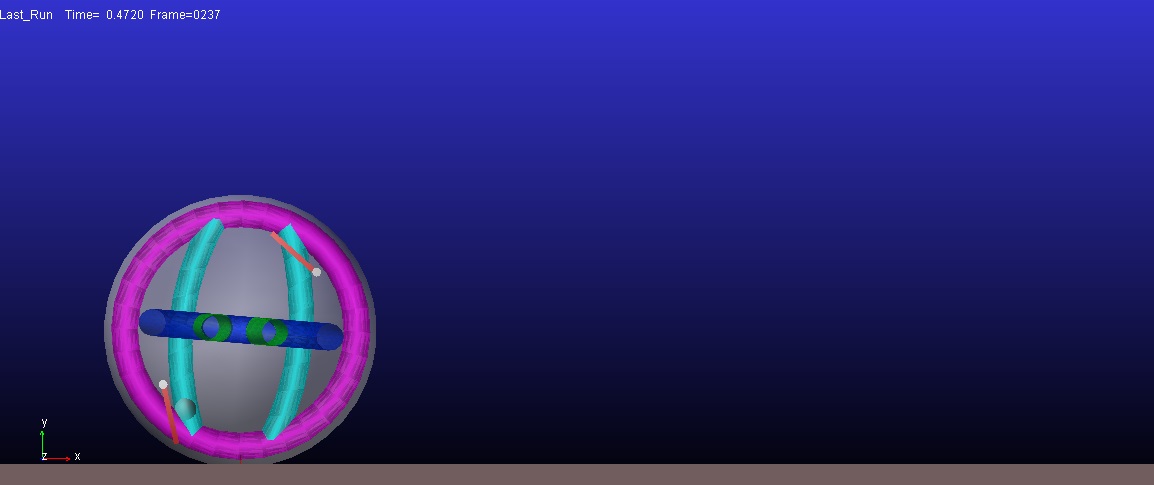} 03
\includegraphics[width=67 mm]{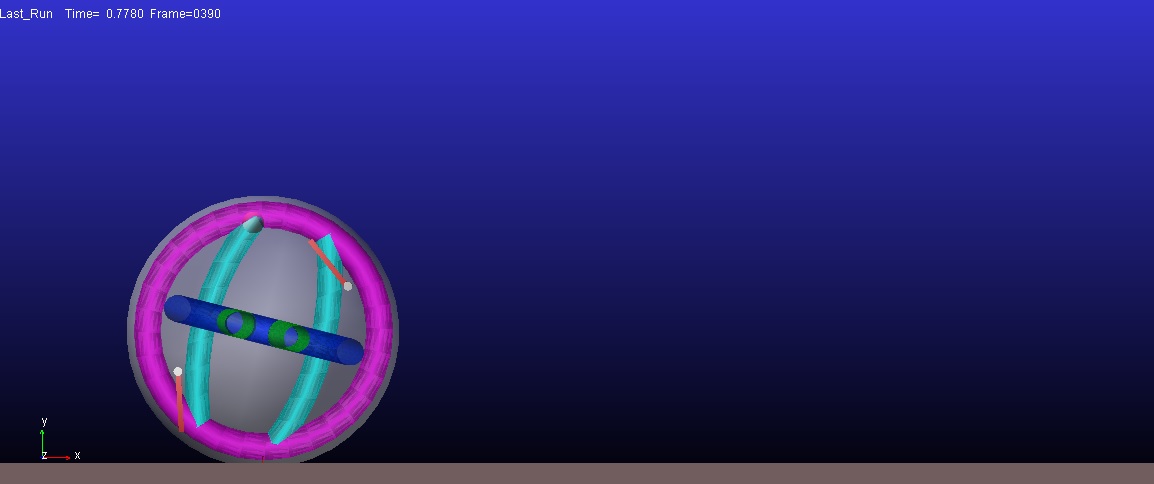} 04
\includegraphics[width=67 mm]{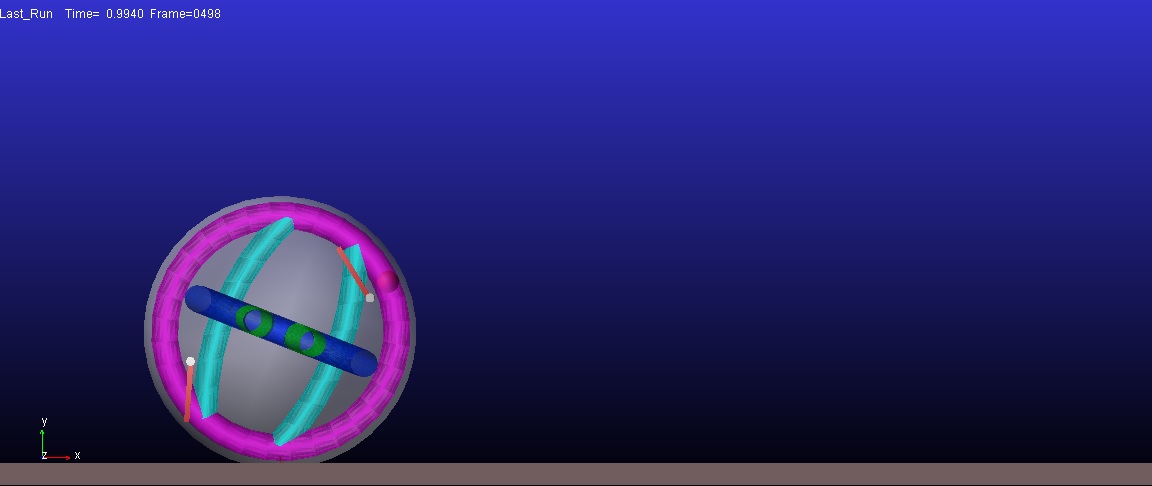} 05
\includegraphics[width=67 mm]{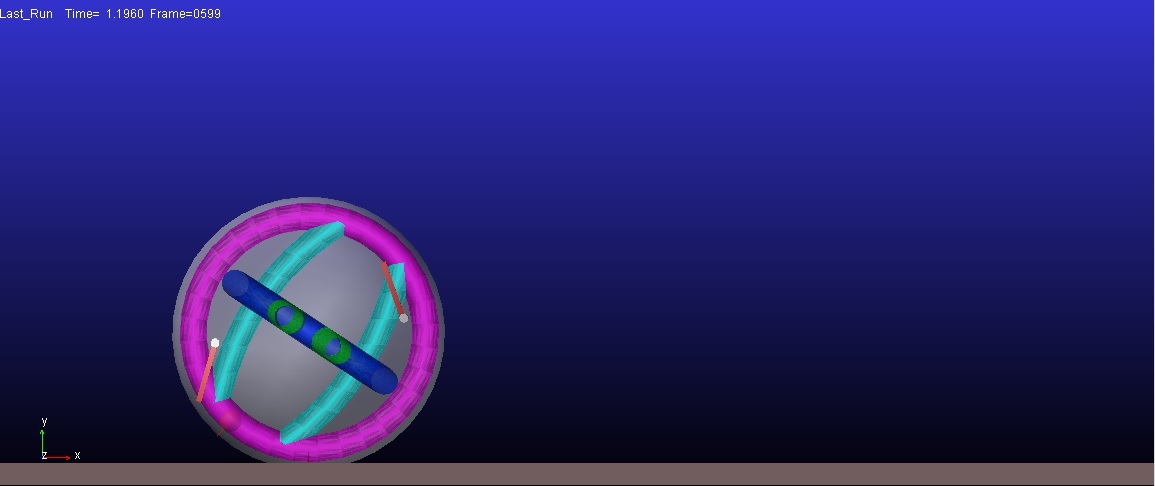} 06
\includegraphics[width=67 mm]{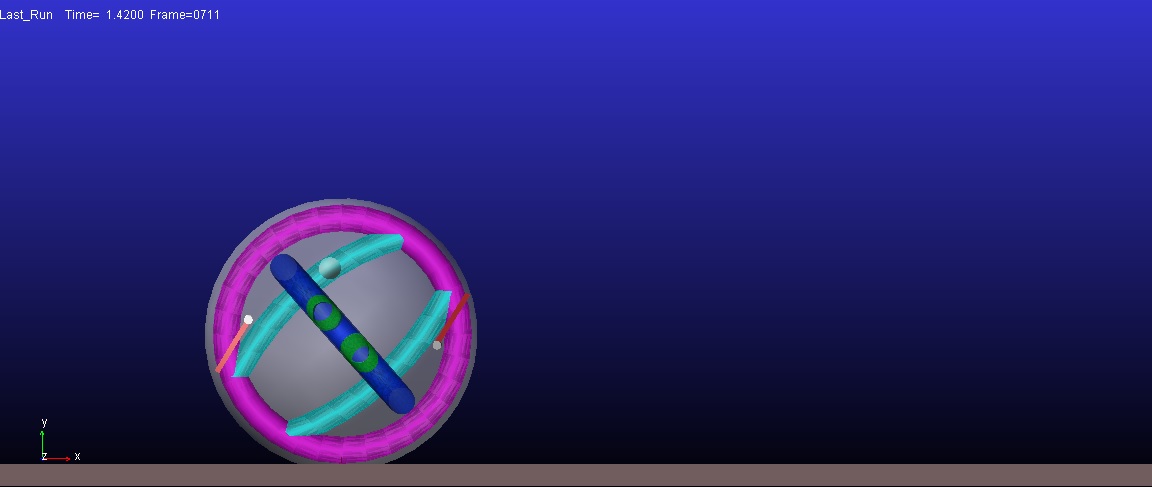} 07
\includegraphics[width=67 mm]{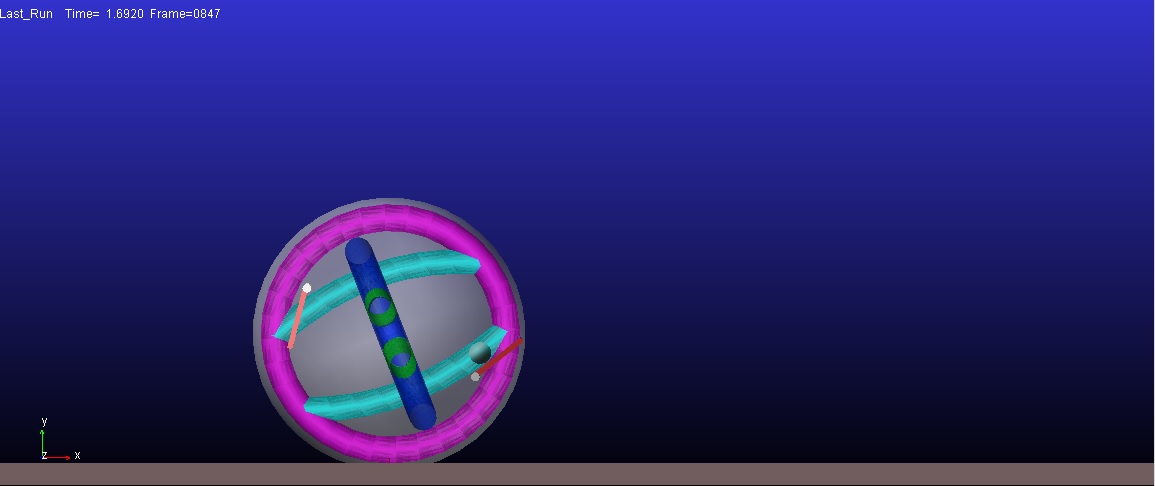} 08
\includegraphics[width=67 mm]{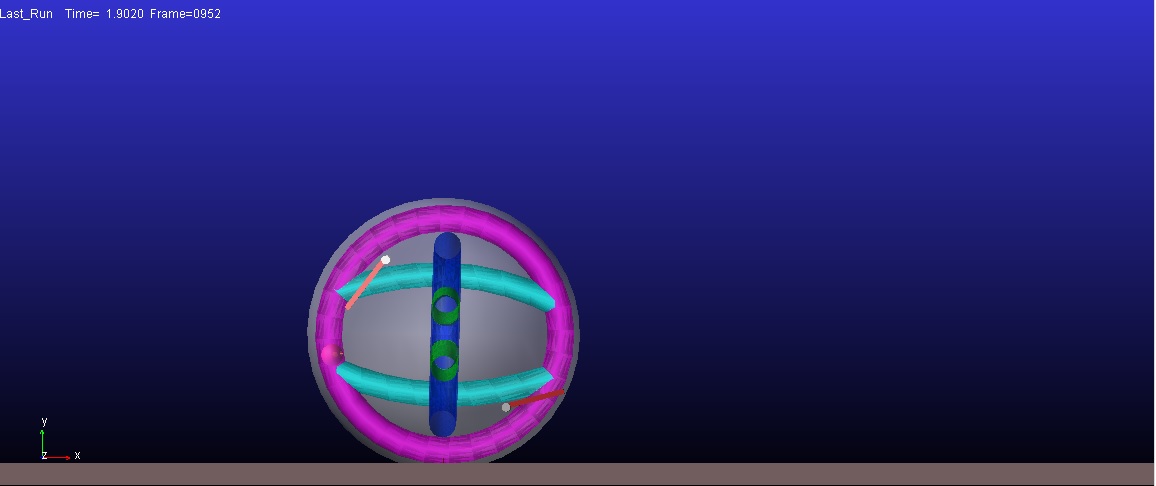} 09
\includegraphics[width=67 mm]{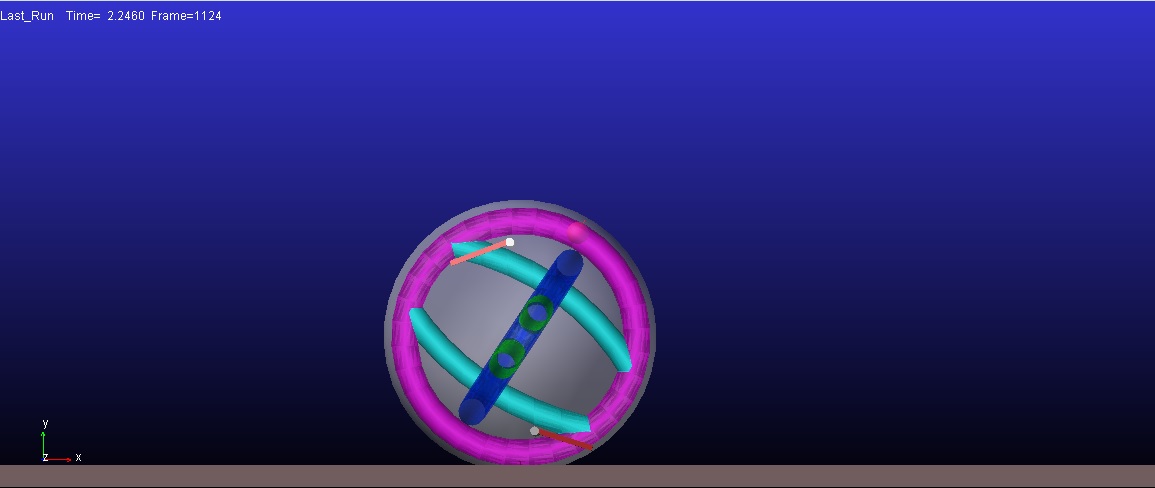} 10
\includegraphics[width=67 mm]{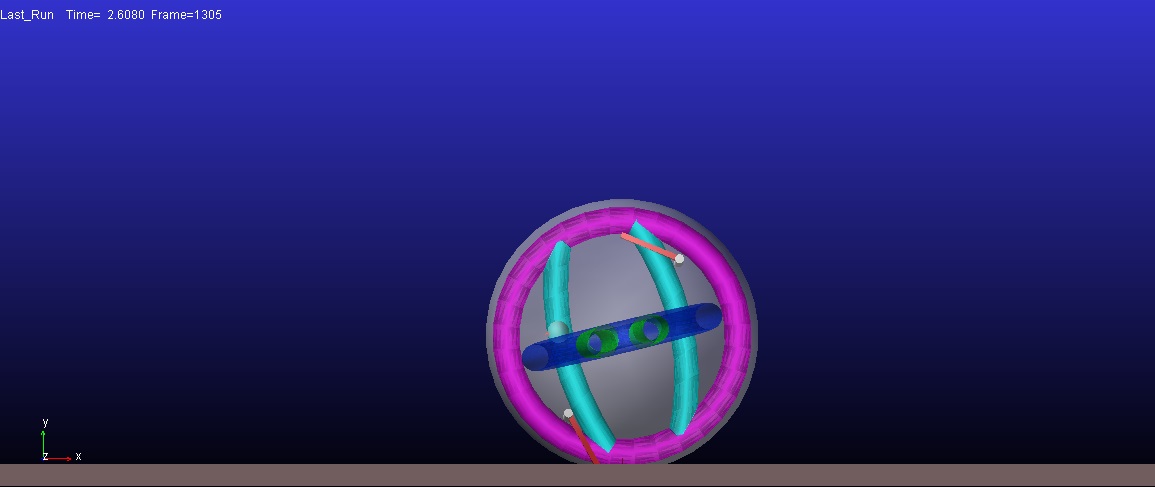} 11
\includegraphics[width=67 mm]{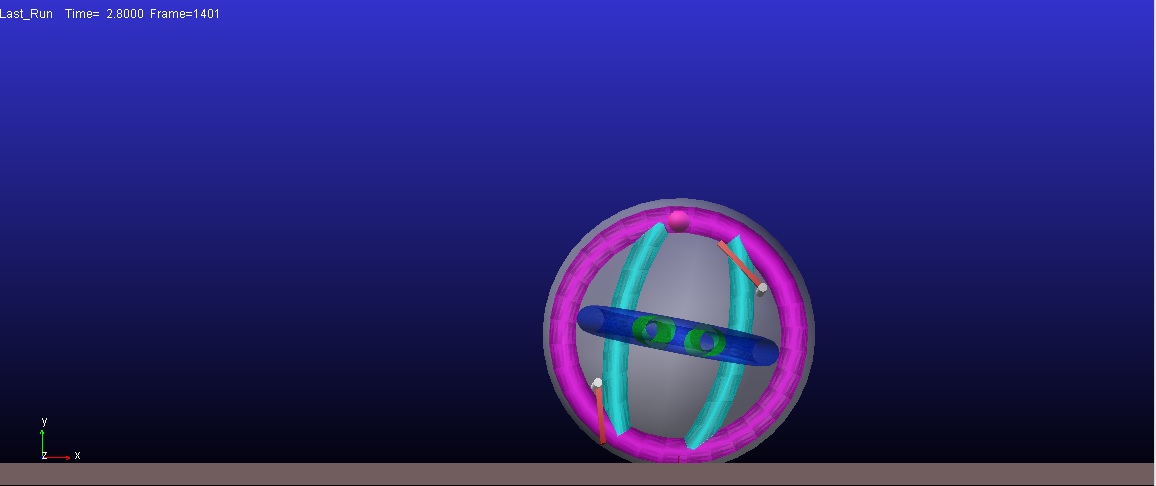} 12
\includegraphics[width=67 mm]{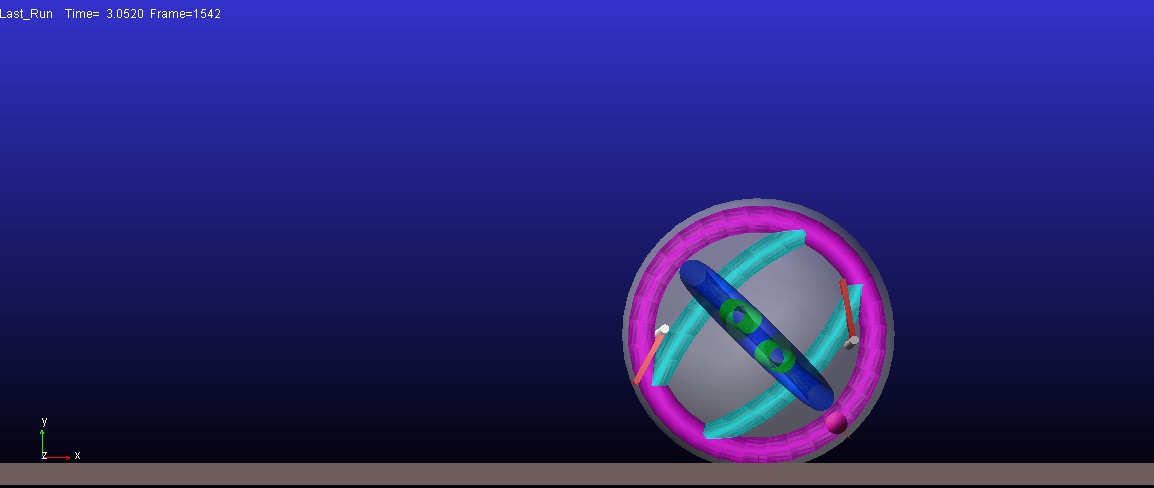} 13 
\includegraphics[width=67 mm]{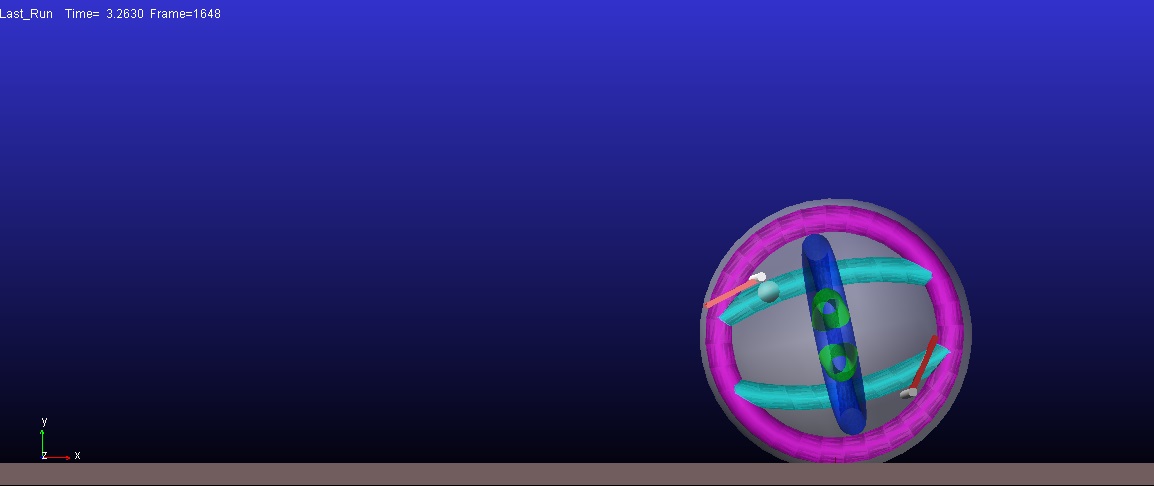} 14

\caption{Sequences of numerical Adams/view simulation with following algorithm motion.}
\end{figure}
After finishing the designing procedure, the robot's simulation played for 3.5 second duration with .001 sampling time. Figure 4.17 demonstrates all steps of algorithms with more detail form during 14 frames interval. After analyzing each step, it can be seen the important movements take place in 7th and 8th step (between 1.3-1.6 sec.). When robot senses that the required angle or movement achieved, it shifted each gates to specific position to prevent any energy losses during motion. The relation of each frame with algorithm in section 2.2.1 is:

Case 'a' $->$ Frame 01;

Case 'b' $->$ Frame 05;

Case 'c' $->$ Frame 06;

Case 'd' $->$ Frame 08;

Case 'e' $->$ Frame 11;

Despite the fact that, The "RollRoller" plotted frames are not exactly continues like proposed algorithm's sequences, neither position sensing nor control feedback were involved to support motion. In other words, algorithm is just for enhancing motion during time in this simulations. For instance, frames 01-04 and 09-10 are presenting that real world factors like friction, damping and other environmental effects can postponed the next algorithm step and robot may keep the same manner of motion algorithm for considerable time duration and then when the controller sense the required position has been achieved, let actuates and gates have next manner.

\begin{figure}[ht!]
\centering
\includegraphics[width=140 mm]{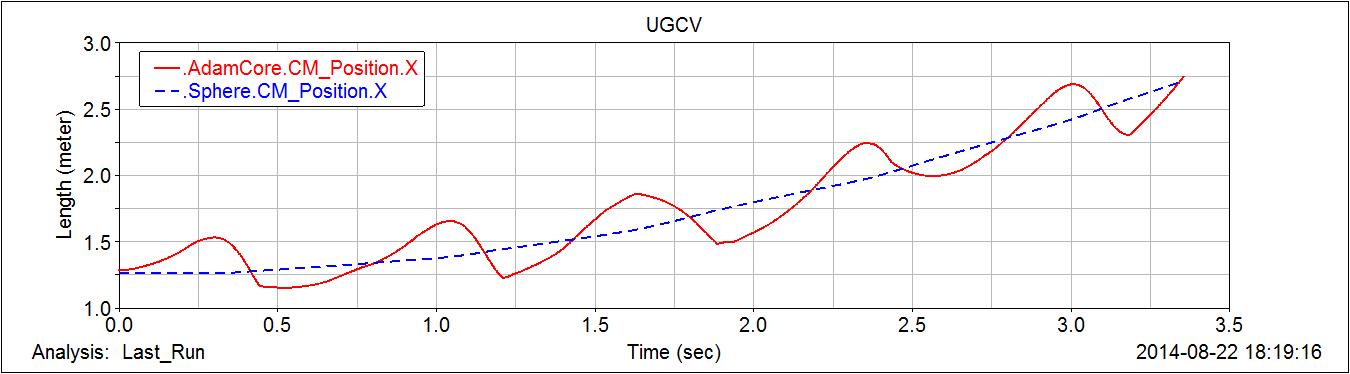} (a)
\includegraphics[width=140 mm]{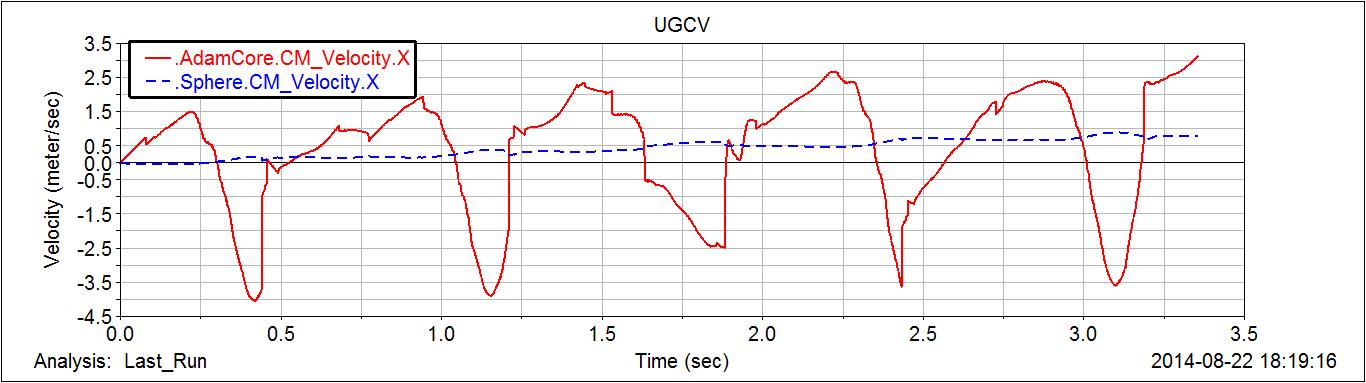} (b)
\caption{(a). Locomotion of "RollRoller" with algorithm (b).Speed of "RollRoller" with algorithm.}
\end{figure}

The output displacement and velocity of core and sphere within 3.3 sec have demonstrated in figure 4.18. The results are completely satisfying and it is showing that most of highly sinusoidal parts of motion specifically in core displacement and velocity have been rectified.Thus, this let sphere have linear locomotion strongly as promised. However, there is small fluctuation in speed of core graph whereabout it is occurring due to type of material chose for pipes and also because core absorbs the force of elliptic-like pendulum force to move so its velocity changes when it enters to GB parts of tubes. And even more the elliptic-like pendulum itself is not perfectly design model, it is just created by connecting series of cylinders discretely. Therefore, upper part of core's velocity in figure can be estimated as rectified.

Figure 4.19 (a) compares both rectified and normal core locomotion. Accordingly , the rectified displacement gives perfect result by comparing the normal method. This phenomenon happens due to leading the core to GB paths to minimized opposite effect motion. Sphere's locomotion in figure 4.19 (b) helps us to realize algorithmic methodology is not remain only in enhancement of core motion but also it damps the spheres motions and considerate the damped energy to raise up the traveled distance. The obtained results validate the strength of "RollRoller" from others, particularly latest developed SMRs locomotion results ( e.g. double pendulum driven robot \cite{Mahboubi} and two internally driven robot \cite{AngularTwo} ) \\

\begin{figure}[ht!]
\centering
\includegraphics[width=140 mm]{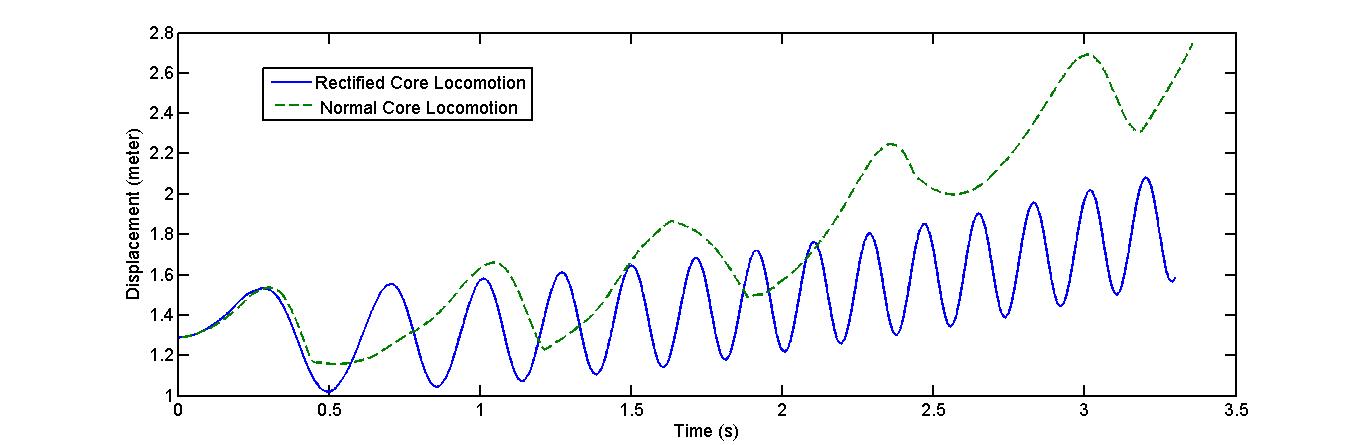} (a)
\includegraphics[width=140 mm]{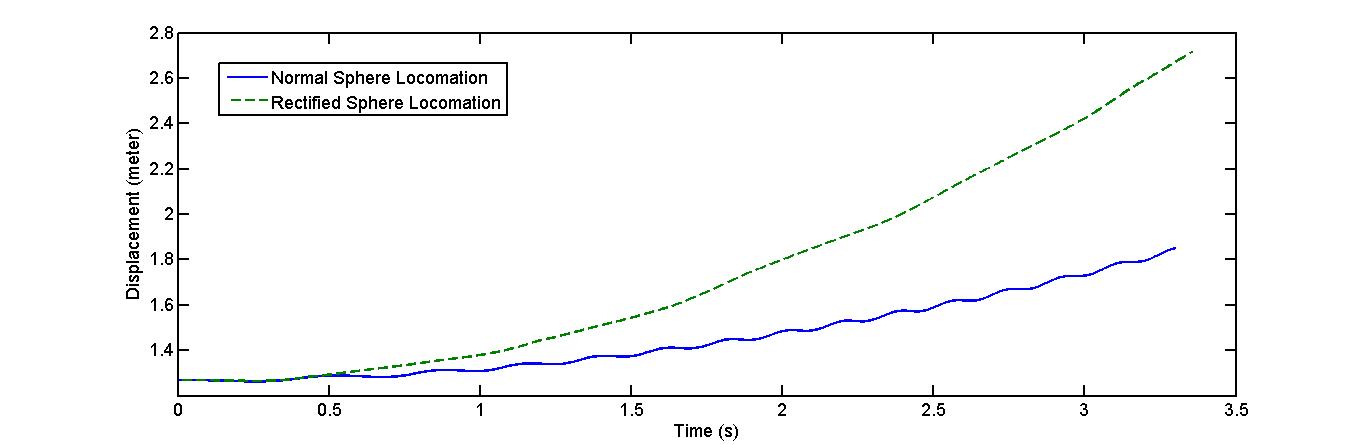} (b)
\caption{(a). Core locomotion comparison (b). Sphere displacement comparison.}
\end{figure}

There are two another serious refinements (figure 4.20), first having ups and downs in Y axis from previous visualization, has completely removed via forward direction algorithm and also it has become more stable and steadier, while from other side the speed of sphere increasing relatively. Second one is the momentum translation which illustrates, as time goes on, SMR starts having more motivation force to jump which may cause serious problem in instability of sphere under normal situation, but from other-side via proposed algorithm the Y axis translational momentum is completely zero. Another vital point should explained about these figures, before .25 sec there is  impulse response in each plot, this happens because total "RollRoller" placed approximately .02 m above ground. After playing simulation, it just falls down short distance and then during the movement it converges to constant value. The reason behind the convergence of Y axis to about .32 meter value relates to plotting of results respect to center of sphere.

\begin{figure}[ht!]
\centering
\includegraphics[width=140 mm]{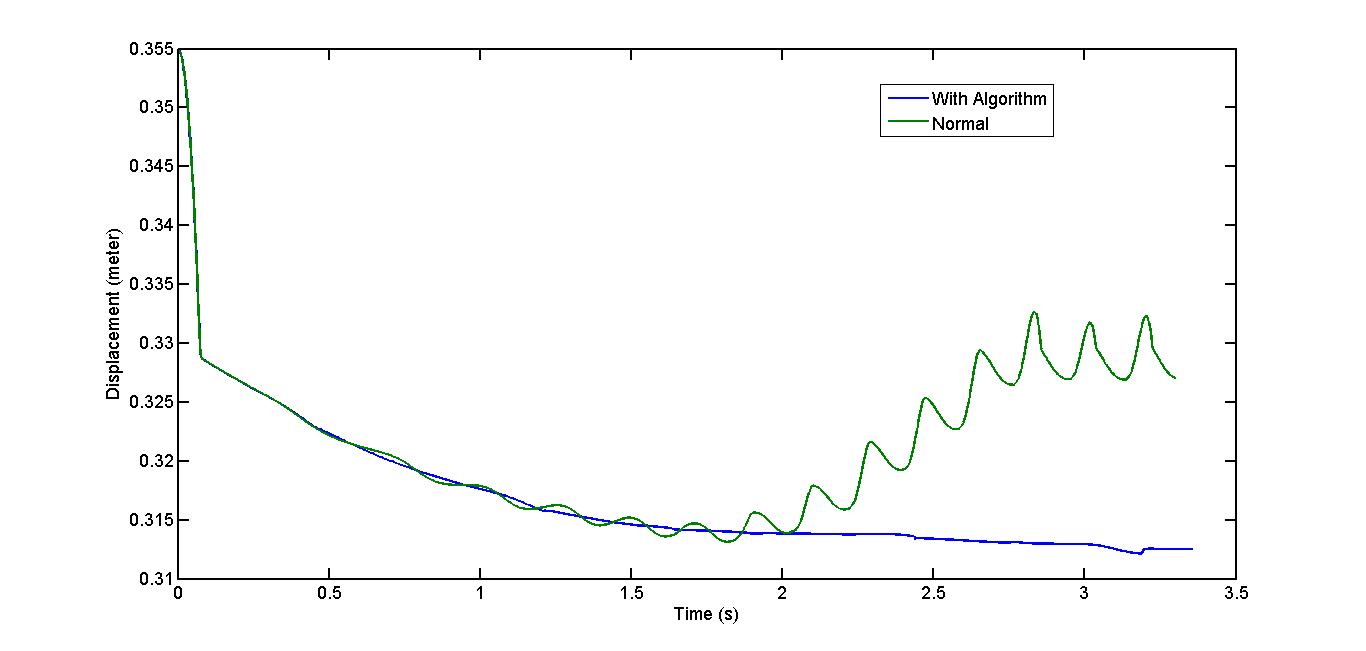} (a)
\includegraphics[width=140 mm]{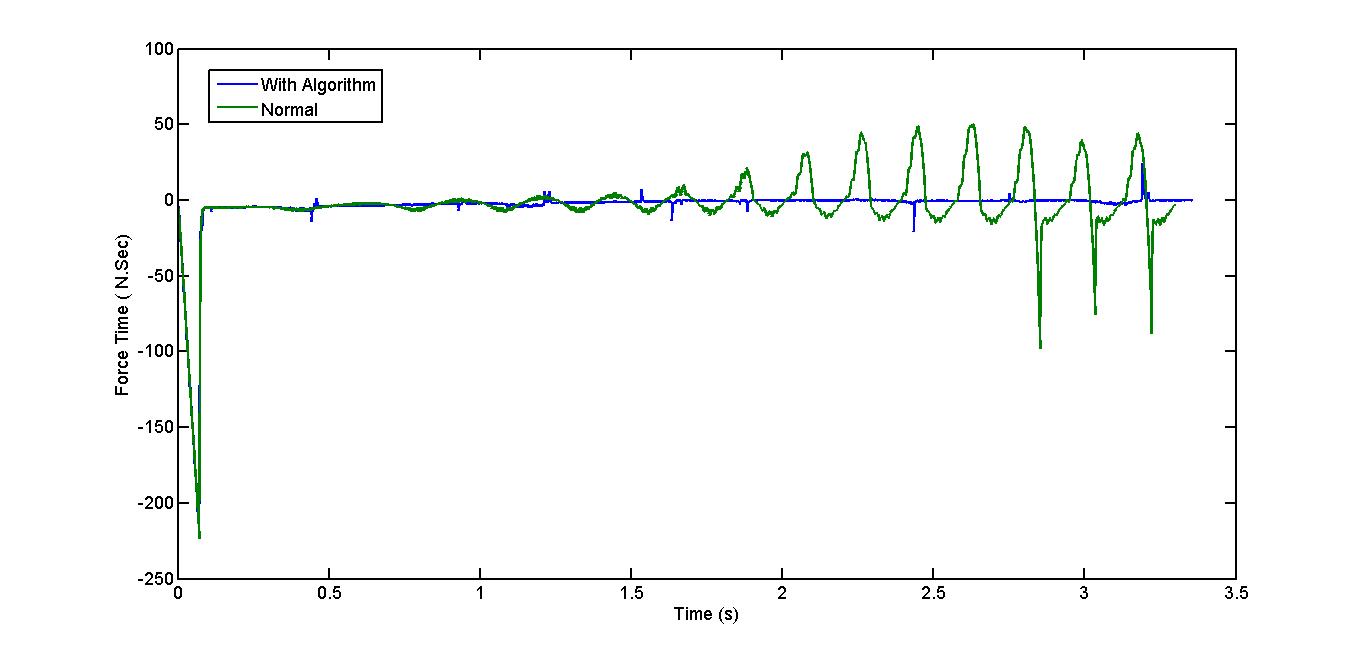} (b)
\caption{(a). Y axis displacement (b). Translational momentum along Y axis.}
\end{figure}

\subsection{Turning Motion}
The circular motion is another ability of this SMR. To give extra detail about robot's morphology during ordinary locomotions this motion also examined as next step. Although, this procedure has to take apart by using the feedback motion because of  nonholonomic nonlinear dynamic, some approximation about basic motion of this SMR is achievable. As same as previous simulation, some enhancements have occurred to Adams/view simulation model as bellow:

\textbf{I)}  Horizontal Tube (HT)'s core was located in right hand side of HT respect to MD as Circular Locomotion Algorithm in section 2.2.2. 

\textbf{II)} Blocker dams was placed around it to prevent core to enter any other pipes (tolerance distance approximation is $^{+}_{-}2cm$).

\textbf{III)}  The required connections between the second core, HT and dams were defined ( Appendix B, figures B.4-5)

The figure 4.21 (a) demonstrates the total structure after creation. The numerical simulation followed path to do the turning has illustrated on figure 4.21 (b). From shown simulation, mass of HT's core keeps its location same in right hand-side corner ( as section 2.2.2 ) while the sphere start turning along Z axis. During this process, VT continue its forward direction motion despite the existence of HT's cores location. This phenomenon cause the center of mass distribution which lead to create two center of mass one located in VT and another in HT. As result by creating specific angular momentum perpendicular to MD , the robot start turning along Y axis. It is better to know, this phenomenon happening without feedback control method. The results are again still satisfying and give accurate 90$^o$ turning but small errors have been appeared because during locomotion HT core swing in its place and prevent locomotion be more clear. The exported data from simulation were recoded in Matlab Program  to see the movement in 3D  for better understanding (figure 4.22).
\begin{figure}[ht!]
\centering
\includegraphics[width=90 mm]{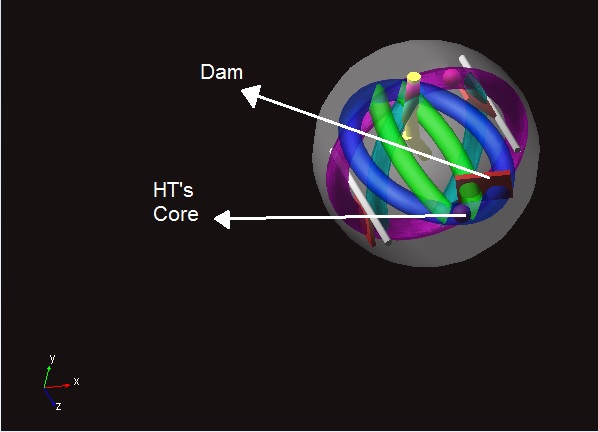} (a)
\includegraphics[width=90 mm]{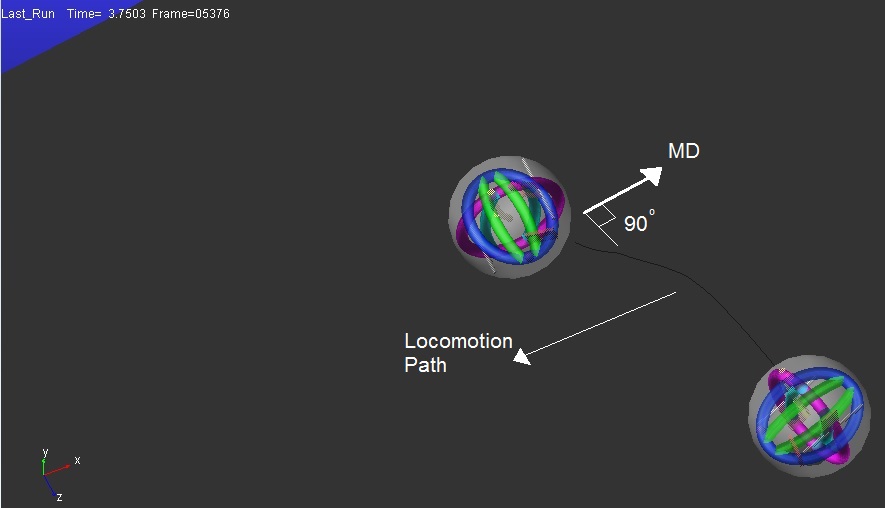} (b)
\caption{(a).The circular locomotion added parts  (b). Circular locomotion coordinates.}
\end{figure}
\begin{figure}[ht!]
\centering
\includegraphics[width=140 mm]{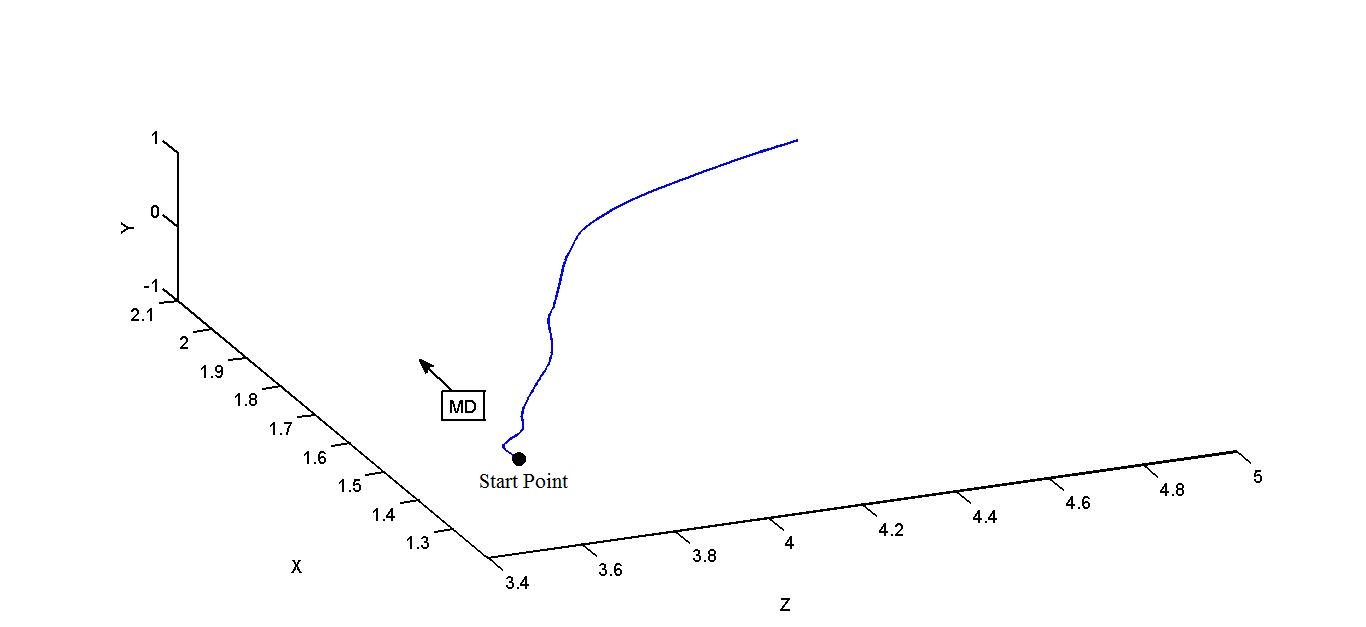} 
\caption{3D movement.}
\end{figure}
\chapter{Conclusion}
Spherical robots have been developing in current years. Their high complexity and hard to manage accurate locomotions were main reasons for not being centralized research subject inside UGV types. In other words, wheeled and legged robots have been more practically active in last years. However, by proposing "RollRoller" with novel and unique IDU mechanism, not only RollRoller can be considered superior from its own kind but also it will be competitive alternative for most of currently using ground mobile robots. As first step, dynamical calculation, including angular momentum and gravitational diversity forces, for this SMR was derived. Furthermore , Numerical simulation on Adams/view program alike Matlab Simulink results validated the mathematical dynamics with achieving same answers. In next step, by modeling the designed motion algorithm in Adams/view, the robots forward direction modified motion represented. Last step was turning locomotion which was simulated successfully. All results were logical even without involving any control methods. This dissertation concludes that about the mathematical basis, "RollRoller" behaving as expected beside strong confirmation from simulations to be highly implementable practical robot.

\chapter{Future Work}
Due to novelty of "RollRoller", there are many cases that will be considered, from making the robot to advancing nonholonomic mathematical analysis and designing suited control methods (i.e. Hybrid control). As first priority "RollRoller" has been designing for implementation to real world including: air producer actuator design, 3D printing of features, embedded system programing and sensor calibration. Additionally, the robot dynamics will be integrated with advance controllers to involve disturbance to system since there is high likelihood of disturbance or exhaustion existence in pressurized air input, even though the sphere's inner body is completely isolated with outer space. Eventually, other advance motions such as angular locomotion and jumping activity will be main subject of future paper publications to let robot utilizes best out of itself. This robot is proposed to be an alternative exploration and rescue mobile robot within these years, however it just requires serious research basis to completely utilize the best performs of it.

\renewcommand{\bibname}{REFERENCES}

\appendix
\chapter{The Matlab Code}
\begin{lstlisting}
%%  Matlab Simulink Constants
g=9.8; %Gavity 
Ms=3; %Sphere Mass
mc=1; %Core Mss
R=0.36; %Sphere Radias
r=.317; %Distance from center of sphere to center of core
Is=(2/3)*Ms*(R^2); %Sphere moment of inertia
Ic=(2/5)*mc*(r^2); % Core moment of inertia
U=1; %Input Torque
b=R*r*mc; % Predefined Constants
h=mc*g*r; %Predefined Constants
D=Ic+(r^2*mc); %Predefined Constants
a=R^2*Ms+Is+Ic+(R^2*mc)+(r^2*mc); %Predefined Constants
%%
\end{lstlisting}
\chapter{Parts Relation In Adams/View Simulation}

\begin{figure}[ht!]
\centering
\includegraphics[width=60 mm]{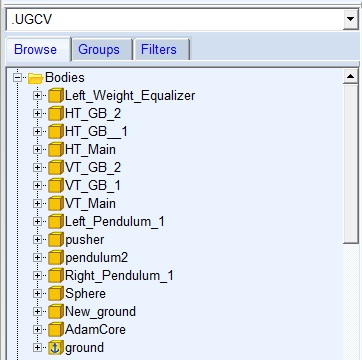}(a)
\includegraphics[width=60 mm]{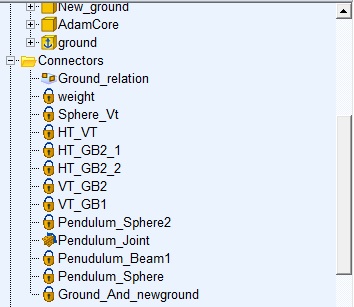}(b)
\includegraphics[width=60 mm]{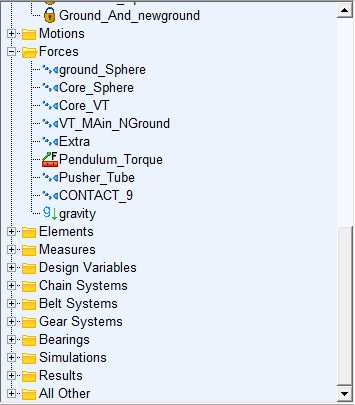}(c)
\caption{(a).Bodies (b).Connectors (c).Forces}
\end{figure}

\begin{figure}[ht!]
\centering
\includegraphics[width=60 mm]{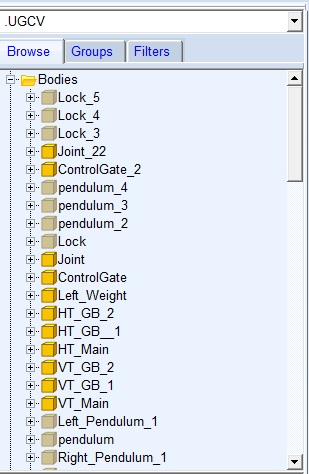}(a)
\includegraphics[width=60 mm]{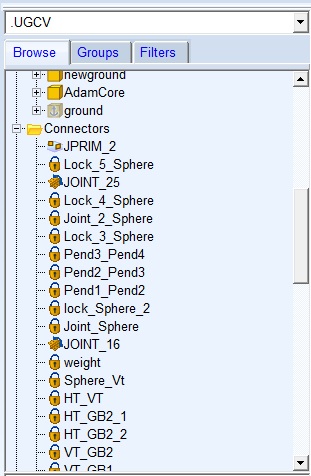}(b)
\includegraphics[width=60 mm]{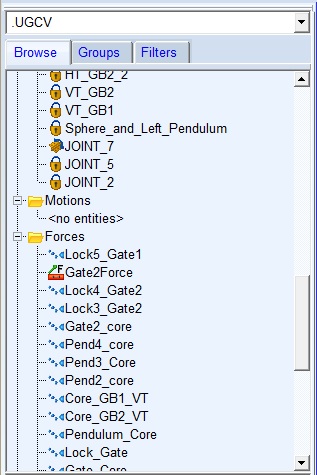}(c)
\includegraphics[width=60 mm]{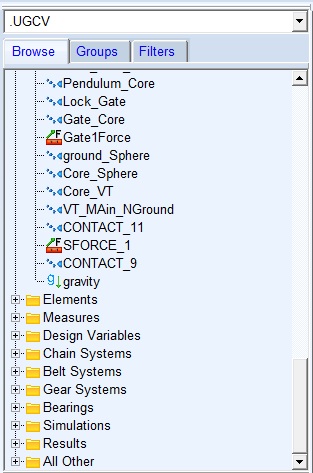}(d)
\caption{Normal Locomotion Definition: (a).Bodies (b).Connectors (c).Forces part 1 (d). Forces part 2}
\end{figure}
\begin{figure}[ht!]
\centering
\includegraphics[width=150 mm]{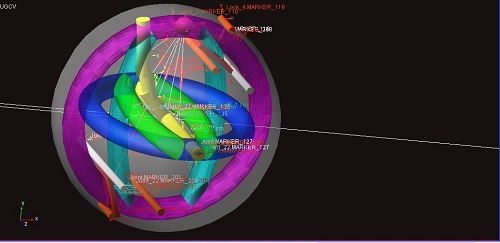}
\caption{The relativities of parts in Adams/view simulation for normal locomotion}
\end{figure}

\begin{figure}[ht!]
\centering
\includegraphics[width=60 mm]{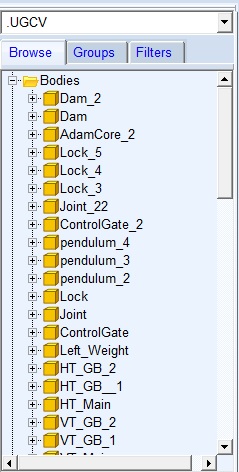}(a)
\includegraphics[width=60 mm]{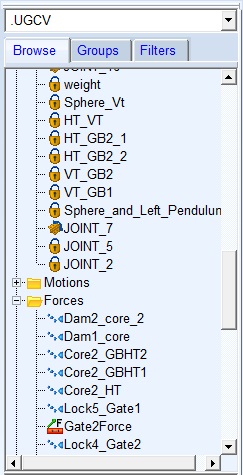}(b)
\includegraphics[width=60 mm]{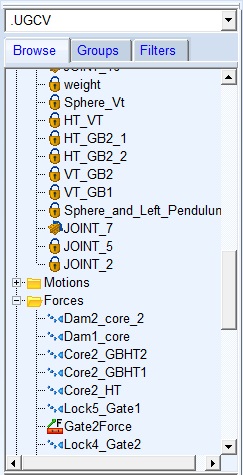}(c)
\includegraphics[width=60 mm]{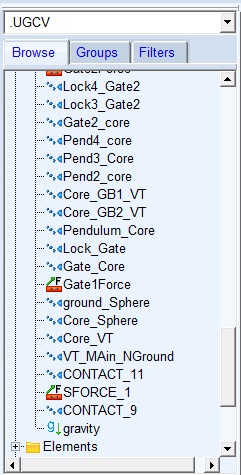}(d)
\caption{Circular Motion Definition: (a).Bodies (b).Connectors (c).Forces part 1 (d). Forces part 2}
\end{figure}

\begin{figure}[ht!]
\centering
\includegraphics[width=150 mm]{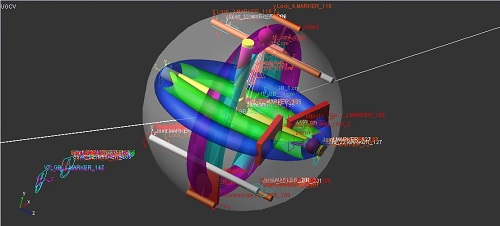}
\caption{The relativities of parts in Adams/view simulation for circular locomotion}
\end{figure}

\end{document}